\pdfoutput=1

\documentclass[11pt]{article}

\usepackage[]{emnlp2021}

\usepackage{times}
\usepackage{latexsym}
\usepackage{svg}
\usepackage[T1]{fontenc}
\usepackage{subcaption}
\usepackage[utf8]{inputenc}
\usepackage{tikz}
\usepackage{pgfplots}
\usepgfplotslibrary{colorbrewer}
\usepackage{microtype}

\usepackage{multirow}
\usepackage{booktabs}
\usepackage{array}
\usepackage{hyperref}

\usepackage{textcomp}  
\usepackage{scalerel}  

\def\stochastic{\scalerel*{\includegraphics{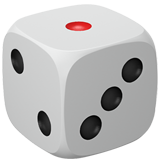}}{\textrm{\textbigcircle}}}
\def\deterministic{\scalerel*{\includegraphics{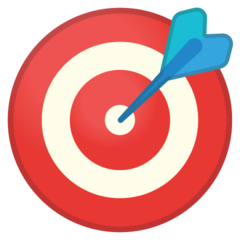}}{\textrm{\textbigcircle}}}

\title{SPECTRA: Sparse Structured Text Rationalization}

\author{Nuno M. Guerreiro$^{1,2}$ \and André F. T. Martins$^{1,2,3}$ \\
$^1$Instituto de Telecomunicações, Lisbon, Portugal \\
$^2$Instituto Superior T\'ecnico \& LUMLIS (Lisbon ELLIS Unit), Lisbon, Portugal\\
$^3$Unbabel, Lisbon, Portugal\\
\footnotesize{\textsf{\{nuno.s.guerreiro, andre.t.martins\}@tecnico.ulisboa.pt}}}

\newcommand{\gen}{\mathsf{gen}}
\newcommand{\pred}{\mathsf{pred}}

\usepackage{amsmath}
\usepackage{amsfonts}
\usepackage{bm}
\usepackage{booktabs}
\usepackage{adjustbox}
\usepackage{tabularx}
\usepackage{pifont}
\usepackage{bbding}

\usepackage{nicefrac}

\definecolor{myblue}{HTML}{0072BC}

\begin{document}
\maketitle
\begin{abstract}
Selective rationalization aims to produce decisions along with  rationales (e.g., text highlights or word alignments between two sentences). 
Commonly, rationales are modeled as stochastic binary masks, requiring sampling-based gradient estimators, which complicates training and requires careful hyperparameter tuning. 
Sparse attention mechanisms are a deterministic alternative, 
but they lack a way to regularize the rationale extraction (e.g., to control the sparsity of a text highlight or the number of  alignments). 
In this paper, we present a unified framework for deterministic extraction of structured explanations via constrained inference on a factor graph, forming a differentiable layer. 
Our approach greatly eases training and rationale regularization,  generally outperforming previous work on what comes to performance and plausibility of the extracted rationales. 
We further provide a comparative study of stochastic and deterministic methods for rationale extraction for classification and natural language inference tasks, jointly assessing their predictive power, quality of the explanations, and model variability.
\end{abstract}

\section{Introduction}

Selective rationalization \citep{lei2016rationalizing, bastings2019interpretable, swanson2020rationalizing} is a powerful explainability method, in which we construct models (\textit{rationalizers}) that produce an explanation or \textit{rationale} (e.g: text highlights or alignments; \citealt{zaidan-eisner-piatko:2007:disc}) along with the decision.

One, if not the main, drawback of rationalizers is that it is difficult to train the generator and the predictor jointly under instance-level supervision \citep{jain2020learning}. Hard attention mechanisms that stochastically sample rationales employ regularization to encourage sparsity and contiguity, and make it necessary to estimate gradients using the score function estimator (SFE), also known as REINFORCE \citep{Williams1992SimpleSG}, or reparameterized gradients \citep{kingma2014autoencoding, jang2017categorical}. Both of these factors substantially complicate training by requiring sophisticated hyperparameter tuning and lead to brittle and fragile models that exhibit high variance over multiple runs.  Other works use strategies such as top-$k$ to map token-level scores to rationales, but also require 
gradient estimations to train both modules jointly \citep{paranjape-etal-2020-information, chang2020invariant}. In turn, sparse attention mechanisms \citep{treviso-martins-2020-explanation} are deterministic and have exact gradients, but 
lack a direct way to control sparsity and contiguity in the rationale extraction. This raises the question: \textit{how can we build an easy-to-train fully differentiable rationalizer that allows for flexible constrained rationale extraction?} 

\begin{table*}[t]
\centering
\footnotesize
\renewcommand\arraystretch{.6}
\begin{tabular}{
>{\arraybackslash}m{3.8cm} >{\arraybackslash}m{2.4cm}
>{\arraybackslash}m{2.6cm}
>{\arraybackslash}m{2.6cm} 
>{\arraybackslash}m{2.5cm}}
\toprule
Method &  Deterministic Training & Exact \newline Gradients & Constrained \newline Extraction & Encourages \newline Contiguity\\ \midrule
\citet{lei2016rationalizing}         & \scriptsize \textcolor{black!70}{\XSolidBrush}  & \scriptsize \textcolor{black!70}{\XSolidBrush}  & \scriptsize \textcolor{black!70}{\XSolidBrush}  & \scriptsize \textcolor{black!100}{\Checkmark}   \\ 
\citet{bastings2019interpretable}      & \scriptsize \textcolor{black!70}{\XSolidBrush}  & \scriptsize \textcolor{black!70}{\XSolidBrush}  & \scriptsize \textcolor{black!100}{\Checkmark} & \scriptsize \textcolor{black!100}{\Checkmark}          \\ 
\citet{treviso-martins-2020-explanation}            & \scriptsize \textcolor{black!100}{\Checkmark} & \scriptsize \textcolor{black!100}{\Checkmark} &   \scriptsize \textcolor{black!70}{\XSolidBrush} & \scriptsize \textcolor{black!70}{\XSolidBrush}               \\
SPECTRA (ours) & \scriptsize \textcolor{black!100}{\Checkmark} & \scriptsize  \textcolor{black!100}{\Checkmark} & \scriptsize \textcolor{black!100}{\Checkmark} & \scriptsize \textcolor{black!100}{\Checkmark} \\ \bottomrule
\end{tabular}
\caption{Positioning of our approach in the literature of rationalization for highlights extraction. Our method is an easy-to-train fully differentiable deterministic rationalizer that allows for flexible rationale regularization.}
\label{Tab:highlightspositioning}
\end{table*}

To answer this question, we introduce \textbf{\underline{sp}ars\underline{e} stru\underline{c}tured \underline{t}ext \underline{ra}tionalization} (\textbf{SPECTRA}), which  employs LP-SparseMAP \citep{niculae2020lpsparsemap}, a constrained structured prediction algorithm, to provide a deterministic, flexible and modular rationale extraction process. We exploit our method's inherent flexibility to extract highlights and interpretable text matchings with a diverse set of constraints. 

Our contributions are:
\begin{itemize}
    \item We present a unified framework for deterministic extraction of  structured rationales (\S\ref{sec:approach}) such as constrained highlights and matchings;
    \item We show how to add constraints on the rationale extraction, and experiment with several structured and hard constraint factors, exhibiting the modularity of our strategy;
    \item We conduct a rigorous comparison between deterministic and stochastic rationalizers (\S\ref{sec:experiments}) for both highlights and matchings extraction.
\end{itemize}

Experiments on selective rationalization for sentiment classification and natural language inference (NLI) tasks show that our proposed approach achieves better or competitive performance and similarity with human rationales, while exhibiting less variability and easing rationale regularization when compared to previous approaches.\footnote{Our library for rationalization is available at \href{https://github.com/deep-spin/spectra-rationalization}{\textsf{https://github.com/deep-spin/spectra-rationalization}}.}

\section{Background}
\label{sec:background}
\subsection{Rationalization for Highlights Extraction}

Rationalization models for highlights extraction, also known as \textit{select-predict} or \textit{explain-predict} models \citep{aligningsocial, Zhang_2021}, are based on a cooperative framework between a rationale generator and a predictor: the generator component encodes the input text and extracts a ``rationale'' (e.g., a subset of highlighted words), and the predictor  classifies the input conditioned only on the extracted rationale. Typically, this is done by obfuscating the words that are not in the rationale with a binary mask.

\paragraph{Highlights Extraction.}  We consider a standard text classification or regression setup, in which we are given an input sequence $\bm{x} \in \mathbb{R}^{D \times L}$, where $D$ is the embedding size and $L$ is the sequence length (number of words), and we want to predict its corresponding label $y \in \mathbb{R}$ for regression or $y  \in \{1, \ldots, C\}$ for classification. A generator model, $\gen$, encodes the input text $\bm{x}$ into token-level scores. Then, a rationale $\bm{z}$, e.g. a binary mask over the tokens, is extracted based on these scores. Subsequently, the predictor model  makes predictions conditioned only on the rationale $\hat{y} = \pred(\bm{z} \odot \bm{x})$, where $\odot$ denotes the Hadamard (elementwise) product.

\paragraph{End-to-end Training and Testing Procedure.} 
While most rationalization methods deterministically select the rationale at test time, there are  differences on how these models are \textbf{trained}. For instance, \citet{lei2016rationalizing} and \citet{bastings2019interpretable} use stochastic binary variables (Bernoulli and HardKuma, respectively), and sample the rationale $\bm{z} \sim \gen(\bm{x}) \in \{0,1\}^L$, whereas \citet{treviso-martins-2020-explanation} make a continuous relaxation of these binary variables and define the rationale as a sparse probability distribution over the tokens, $\bm{z} = \mathsf{sparsemax}(\gen(\bm{x}))$ or $\bm{z} = \alpha\text{-}\mathsf{entmax}(\gen(\bm{x}))$. In the latter approach, instead of a binary vector, we have $\bm{z} \in \triangle^{L-1}$, where $\triangle^{L-1}$ is the $L-1$ probability simplex $\triangle^{L-1} := \{\bm{p} \in \mathbb{R}^L: \bm{1}^\top \bm{p} = 1, \bm{p} \geq 0 \}$. Words receiving non-zero probability are considered part of the rationale. 

Rationalizers that use hard attention mechanisms or heuristics to extract the rationales are distinctively hard to train end-to-end, as they require marginalization over all possible rationales, which is intractable in practice. Thus, recourse to sampling-based gradient estimations is a necessity, either via REINFORCE-style training, which exhibits high variance \citep{lei2016rationalizing, chang2020invariant}, or via reparameterized gradients \citep{bastings2019interpretable, paranjape-etal-2020-information}. This renders training these models a complex and cumbersome task. These approaches are often brittle and fragile for the high sensitivity that they show to changes in the hyperparameters and to variability due to sampling. On the other hand, existing rationalizers that use sparse attention mechanisms \citep{treviso-martins-2020-explanation} such as sparsemax attention, while being deterministic and end-to-end differentiable, do not have a direct handle to constrain the rationale in terms of sparsity and contiguity. We endow them with these capabilities in this paper as shown in Table \ref{Tab:highlightspositioning}, where we position our work in the literature for highlights extraction.

\paragraph{Constrained Rationale Extraction.} Existing rationalizers are \emph{extractive}: they select and extract words or word pairs to form the rationale. Since a rationalizer that extracts the whole input would be meaningless as an explainer, they must have a length constraint or a sparsity inducing component. Moreover, rationales are idealized to encourage selection of contiguous words, as there is some evidence that this improves readibility \citep{jain2020learning}. Some works opt to introduce regularization terms placed on the binary mask such as the $\ell_1$ norm and the fused-lasso penalty to encourage sparse and compact rationales \citep{lei2016rationalizing, bastings2019interpretable}. Others use hard constraints through heuristics such as top-$k$, which is not contiguous but sparse, or select a chunk of text with a pre-specified length that corresponds to the highest total score over all possible spans of that length \cite{chang2020invariant, paranjape-etal-2020-information, jain2020learning}. Sparse attention mechanisms can also be used to extract rationales, but since the rationales are constrained to be in the simplex, controlling the number of selected tokens and simultaneously promoting contiguity is non-trivial.

\subsection{Rationalization for Matchings Extraction}
For this task, we consider a natural language inference setup in which classification is made based on two input sentences: a premise $\bm{x}_P \in \mathbb{R}^{D \times L_P}$ and a hypothesis $\bm{x}_H \in \mathbb{R}^{D \times L_H}$, where $L_P$ and $L_H$ are the sequence lengths of the premise and hypothesis, respectively, and $D$ is the embedding size. A generator model ($\gen$) encodes  $\bm{x}_P$ and $\bm{x}_H$ separately and then computes pairwise costs between the encoded representations to produce a score matrix $\bm{S} \in \mathbb{R}^{L_P \times L_H}$. The score matrix $\bm{S}$ is then used to compute an alignment matrix $\bm{Z} \in \mathbb{R}^{L_P \times L_H}$, where $z_{ij}=1$ if the $i\textsuperscript{th}$ premise word is aligned to the $j\textsuperscript{th}$ word in the hypothesis. $\bm{Z}$ subsequently acts as a sparse mask to obtain text representations that are  aggregated with the original encoded sequences and fed to a predictor to obtain the output predictions.

\subsection{Structured Prediction on Factor Graphs}
\label{background:lpsparsemap}

Finding the highest scored rationale under the constraints described above is a structured prediction problem, which involves searching over a very large and combinatorial space. 
We assume that a rationale $\bm{z}$ can be represented as an $L$-dimensional binary vector. 
For example, in highlights extraction, $L$ is the number of words in the document and $\bm{z}$ is a binary mask selecting the relevant words; and in the extraction of matchings, $L = L_P\times L_H$ and $\bm{z}$ is a flattened binary vector whose entries indicate if a premise word is aligned to a word in the hypothesis.  
We let $\mathcal{Z} \subseteq \{0,1\}^{L}$ be the set of rationales that satisfy the given constraints, and let $\bm{s} = \gen(\bm{x}) \in \mathbb{R}^L$ be a vector of scores. 

\paragraph{Factor Graph.} In the sequel, we consider problems that consist of multiple interacting subproblems. \citet{niculae2020lpsparsemap} present structured differentiable layers, which decompose a given problem into simpler  subproblems, instantiated as local factors that  must agree when overlapped. 
Formally, we assume a factor graph $\mathcal{F}$, where each factor $f \in \mathcal{F}$ corresponds to a subset of variables. We denote by $\bm{z}_f = (\bm{z}_i)_{i \in f}$ the vector of variables corresponding to factor $f$. 
Each factor has a local score function $h_f(\bm{z}_f)$. 
Examples are \textbf{hard constraint factors}, which take the form
\begin{equation}
    h_f(\bm{z}_f) = \left\{
    \begin{array}{ll}
        0 & \text{if $\bm{z}_f \in \mathcal{Z}_f$} \\
        -\infty & \text{otherwise}, 
    \end{array}
    \right.
\end{equation}
where $\mathcal{Z}_f$ is a polyhedral set imposing hard constraints (see Table \ref{Tab:constraints} for examples); and \textbf{structured factors}, which define more complex functions with structural dependencies on $\bm{z}_f$, such as
\begin{equation}
    h_f(\bm{z}_f) = \sum_{i=1}^{L-1} r_{i, i+1} z_{i, i+1},
\end{equation}
where $r_{i, i+1} \in \mathbb{R}$ are edge scores, 
which together define a \textbf{sequential factor}.
We require that for any factor the following local subproblem is tractable:
\begin{equation}
    \hat{\bm{z}}_f = \arg\max_{\bm{z}_f \in \{0,1\}^{|f|}} \bm{s}_f^\top \bm{z}_f + h_f(\bm{z}_f). 
\end{equation}

\paragraph{MAP inference.} The problem of identifying the highest-scoring global structure, known as \textbf{maximum \textit{a posteriori}} (MAP) \textbf{inference}, is written as:
\begin{equation}\label{eq:map}
    \hat{\bm{z}} = \arg\max_{\bm{z} \in \{0,1\}^L} \underbrace{\bigl( \bm{s}^\top \bm{z} + \sum_{f \in \mathcal{F}} h_f(\bm{z}_f) \bigr)}_{\mathrm{score}(\bm{z}; \bm{s})}.
\end{equation}
The objective being maximized is the global score function $\mathrm{score}(\bm{z}; \bm{s})$, which combines information coming from all factors. The solution of the MAP problem is a vector $\hat{\bm{z}}$ whose entries are zeros and ones. However, it is often difficult to obtain an exact maximization algorithm for complex structured problems that involve interacting subproblems that impose global agreement constraints.

\paragraph{Gibbs distribution and sampling.} The global score function can be used to define a Gibbs distribution $p(\bm{z}; \bm{s}) \propto \exp(\mathrm{score}(\bm{z}; \bm{s}))$. The MAP in \eqref{eq:map} is the mode of this distribution. 
Sometimes (e.g. in stochastic rationalizers) we want to sample from this distribution, $\hat{\bm{z}} \sim p(\bm{z}; \bm{s})$. Exact, unbiased samples are often intractable to obtain, and approximate sampling strategies have to be used, such as perturb-and-MAP  \citep{perturbandMAP, corro2019differentiable, corro2019learning}. These strategies necessitate gradient estimators for end-to-end training, which are often obtained via REINFORCE \citep{Williams1992SimpleSG} or reparametrized gradients \citep{kingma2014autoencoding, jang2017categorical}.

\paragraph{LP-MAP inference.} 
In many cases, the MAP problem \eqref{eq:map} is intractable due to the overlapping interaction of the factors $f \in \mathcal{F}$. 
A commonly used relaxation is to replace the integer constraints $\bm{z} \in \{0,1\}^L$ by continuous constraints, leading to:
\begin{equation}\label{eq:lpmap}
    \hat{\bm{z}} = \arg\max_{\bm{z} \in \color{myblue}{[0,1]^L}} \mathrm{score}(\bm{z}; \bm{s}).
\end{equation}
The problem above is known as LP-MAP inference \citep{wainwright2008graphical}. In some cases (for example, when the factor graph $\mathcal{F}$ does not have cycles), LP-MAP inference is \emph{exact}, i.e., it gives the same results as MAP inference. In general, this does not happen, but for many problems in NLP, LP-MAP relaxations are often nearly optimal \citep{koo-etal-2010-dual, martins2015ad3}. Importantly, computation in the hidden layer of these problems may render the network unsuitable for gradient-based training, as with MAP inference.

\paragraph{LP-SparseMAP inference.} The optimization problem respective to LP-SparseMAP is the $\ell_2$ regularized LP-MAP \cite{niculae2020lpsparsemap}:
\begin{equation}\label{eq:lpsparsemap}
    \hat{\bm{z}} = \arg\max_{\bm{z} \in \color{myblue}{[0,1]^L}} \bigl( \mathrm{score}(\bm{z}; \bm{s}) {\color{myblue}{- \nicefrac{1}{2}\|\bm{z}\|^2}} \bigr).
\end{equation}
Unlike MAP and LP-MAP, the LP-SparseMAP relaxation is suitable to train with gradient backpropagation. Moreover, it favors sparse vectors $\hat{\bm{z}}$, i.e., vectors that have only a few non-zero entries. One of the most appealing features of this method is that it is modular: an arbitrary complex factor graph can be instantiated as long as a MAP oracle for each of the constituting factors is provided. This approach generalizes SparseMAP \citep{niculae2018sparsemap}, which requires an exact MAP oracle for the factor graph in its entirety. In fact, LP-SparseMAP  recovers SparseMAP when there is a single factor $\mathcal{F} = \{f\}$. By only requiring a MAP oracle for each $f \in \mathcal{F}$, LP-SparseMAP makes it possible to instantiate more expressive factor graphs for which MAP is typically intractable. Table \ref{Tab:constraints} lists several logic constraint factors which are used in this paper.


\begin{table}[t]
\centering
\footnotesize
\renewcommand\arraystretch{1}
\begin{tabularx}{0.8\columnwidth}{p{0.3\columnwidth} p{0.4\columnwidth}}
\toprule
Factor Name &  Imposed Constraint \\ \midrule
$\mathsf{XOR}$         &  $\sum z_k = 1$                  \\ 
$\mathsf{AtMostOne}$ & $\sum z_k \leq 1$                   \\ 
$\mathsf{BUDGET}$            &  $\sum z_k \leq B$                  \\
\bottomrule
\end{tabularx}
\caption{Collection of logic factors and its imposed constraints. Each of these factors defines a constraint set $\mathcal{Z}_f$ as described in \S\ref{background:lpsparsemap}.}
\label{Tab:constraints}
\end{table}

\section{Deterministic Structured Rationalizers}
\label{sec:approach}

The idea behind our approach for selective rationalization is very simple: leverage the inherent flexibility and modularity of LP-SparseMAP for constrained, deterministic and fully differentiable rationale extraction.

\subsection{Highlights Extraction}

\paragraph{Model Architecture.} We use the model setting described in \S\ref{sec:background}. First, a generator model produces token-level scores $s_i, i \in \{1, \dots, L\}$. We propose replacing the current rationale extraction mechanisms (e.g. sampling from a Bernoulli distribution, or using sparse attention mechanisms) with an LP-SparseMAP extraction layer that computes token-level values $\hat{\bm{z}} \in [0,1]^L$, which are then used to mask the original sequence for prediction. Due to LP-SparseMAP's propensity for sparsity, many entries in $\hat{\bm{z}}$ will be zero, which approaches what is expected from a binary mask.

\paragraph{Factor Graphs.} The definition of the factor graph $\mathcal{F}$ is central to the rationale extraction, as each of the local factors $f \in \mathcal{F}$ will impose constraints on the highlight. We start by instantiating a factor graph with $L$ binary variables (one for each token) and a pairwise factor for every pair of contiguous tokens:
\begin{align}
    \mathcal{F} = \{{\mathsf{PAIR}}(&z_i, z_{i+1}; r_{i, i+1}): 1 \leq i < L\},
\end{align}
which yields the binary pairwise MRF (\S\ref{background:lpsparsemap}) \begin{align}
\mathrm{score}(\bm{z}; \bm{s}) = 
\sum_{i=1}^L s_i z_i + \sum_{i=1}^{L-1} r_{i, i+1} z_i z_{i+1}.
\end{align}
Instantiating this factor with non-negative edge scores, $r_{i, i+1} \geq 0$, encourages contiguity on the rationale extraction. Making use of the modularity of the method, we impose sparsity by further adding a $\mathsf{BUDGET}$ factor (see Table \ref{Tab:constraints}):
{\begin{align}
    \mathcal{F} = &\{{\mathsf{PAIR}}(z_i, z_{i+1}; r_{i, i+1}): 1 \leq i < L\}\nonumber \\
    &\cup\, \{\mathsf{BUDGET}(z_1, \dots, z_L; B)\}.
    \label{eq: factorpair}
\end{align}
The size of the rationale is constrained to be, at most, $B\%$ of the input document size. Intuitively, the lower the $B$, the shorter the extracted rationales will be. Notice that this graph is composed of $L$ local factors. Thus, LP-SparseMAP would have to enforce agreement between all these factors in order to compute $\bm{z}$. Interestingly, factor graph representations are usually not unique. In our work, we instantiate an equivalent formulation of the factor graph in Eq. \ref{eq: factorpair} that consists of a single factor, \textbf{\textsf{H:SeqBudget}}. This factor can be seen as an extension of that of the LP-Sequence model in \citet{niculae2020lpsparsemap}: a linear-chain Markov factor with MAP provided by the Viterbi algorithm \citep{viterbi_orig, Viterbi}. The difference resides in the additional budget constraints that are incorporated in the MAP decoding. These constraints can be handled by augmenting the number of states in the dynamic program to incorporate how many words in the budget have already been consumed at each time step, leading to time complexity $\mathcal{O}(LB)$.

\subsection{Matchings Extraction}
\label{sec:matchings-extraction}

\paragraph{Model Architecture.}
Our architecture is inspired by ESIM \citep{chen-etal-2017-enhanced}. First, a generator model encodes two documents $\bm{x}_P$, $\bm{x}_H$ separately to obtain the encodings $(\Tilde{\bm{h}}_1^P, \dots, \Tilde{\bm{h}}_{L_P}^P)$ and $(\Tilde{\bm{h}}_1^H, \dots, \Tilde{\bm{h}}_{L_H}^H)$, respectively. Then, we compute alignment dot-product pairwise scores between the encoded representations to produce a score matrix $\bm{S} \in \mathbb{R}^{L_P \times L_H}$ such that $s_{ij} = \langle\Tilde{\bm{h}}_i^P, \Tilde{\bm{h}}_j^H\rangle$. We  use LP-SparseMAP to obtain $\bm{Z}$, a constrained structured symmetrical alignment $\bm{Z}$ in which $z_{ij} \in [0,1]$, as described later. Then, we ``augment'' each word in the premise and hypothesis with the corresponding aligned weighted average by computing $\Bar{\bm{h}}_i^P = \left[\Tilde{\bm{h}}_i^P, \sum_{j} z_{ij}\, \Tilde{\bm{h}}_j^H\right]$ and $\Bar{\bm{h}}_j^H = \left[\Tilde{\bm{h}}_j^H, \sum_{i} z_{ji}\, \Tilde{\bm{h}}_i^P\right]$, and separately feed these vectors to another encoder and pool to find representations $\bm{r}^P$ and $\bm{r}^H$. Finally, the feature vector $\bm{r} = [\bm{r}^P, \bm{r}^H, \bm{r}^P-\bm{r}^H, \bm{r}^P \odot \bm{r}^H]$ is fed to a classification head for the final prediction. We also experiment with a strategy in which we assume that the hypothesis is known and the premise is masked for \textit{faithful} prediction. We consider  $\Bar{\bm{h}}_i^P = \left[\sum_{j} z_{ij}\, \Tilde{\bm{h}}_j^H\right]$, such that the only information about the premise that the model has to make a prediction comes from the alignment and its masking of the encoded representation.

\paragraph{Factor Graphs.} We instantiate three different factor graphs for matchings extraction. The first -- \textbf{\textsf{M:XorAtMostOne}} -- is the same as the LP-Matching factor used in \citet{niculae2020lpsparsemap} with one \textsf{XOR} factor per row and one \textsf{AtMostOne} factor per column: 
{\begin{align}
    \mathcal{F} =\, &\{{\mathsf{XOR}}(z_{i1}, ... ,  z_{in}): 1 \leq i \leq L_P \\
    &\cup \{\mathsf{AtMostOne}(z_{1j}, ... , z_{mj}): 1 \leq j \leq L_H\} \nonumber
    \label{eq: xoratmostone}
\end{align}
which requires at least one active alignment for each word of the premise, since the $i^{\textrm{th}}$ word in the premise \textbf{must} be connected to the hypothesis. The $j^{\textrm{th}}$ word in the hypothesis, however, is not constrained to be aligned to any word in the premise.  In the second factor graph -- \textbf{\textsf{M:AtMostOne2}} -- we alleviate the \textsf{XOR} restriction on the premise words to an $\mathsf{AtMostOne}$ restriction. The expected output is a sparser matching for there is no requirement of an active alignment for each word of the premise. The third factor graph -- \textsf{\textbf{M:Budget}} -- allows us to have more refined control on the sparsity of the resulting matching, by adding an extra global $\mathsf{BUDGET}$ factor (with budget $B$) to the factor graph of {\textsf{M:AtMostOne2}} so that the resulting matching will have at most $B$ active alignments.

\paragraph{Stochastic Matchings Extraction.} Prior work for selective rationalization of text matching uses constrained variants of optimal transport to obtain the rationale \citep{swanson2020rationalizing}. Their model is end-to-end differentiable using the Sinkhorn algorithm \citep{sinkhorn}. Thus, in order to provide a comparative study of stochastic and deterministic methods for rationalization of text matchings, we implement a perturb-and-MAP rationalizer (\S \ref{background:lpsparsemap}). We perturb the scores $s_{ij}$ by computing $\bm{\Tilde{S}} = \bm{S} + \bm{P}$, in which each element of $\bm{P}$ contains random samples from the Gumbel distribution, $p_{ij} \sim \mathcal{G}(0,1)$. We utilize these perturbed scores to compute non-symmetrical alignments from the premise to the hypothesis and vice-versa, such that their entries are in $[0,1]$. At test time, we obtain the most probable matchings, such that their entries are in $\{0,1\}$. These matchings are such that every word in the premise \textbf{must} be connected to a single word in the hypothesis and vice-versa.

\section{Experimental Setup}\label{sec:experiments}
\subsection{Highlights for Sentiment Classification}
\paragraph{Data and Evaluation.} We used the SST, AgNews, IMDB, and Hotels datasets for text classification and the BeerAdvocate dataset for regression. The statistics and details of all datasets can be found in \S\ref{sec:data_highlights}. The rationale specified lengths, as percentage of each document, for the strategies that impose fixed sparsity are 20\% for the SST, AgNews and IMDB datasets, 15\% for the Hotels dataset, and 10\% for the BeerAdvocate dataset.  We evaluate end task performance (Macro $F_1$ for classification tasks and MSE for regression), and matching with human annotations through token-level $F_1$ score \citep{deyoung2019eraser} for the datasets that contain human annotations.

\paragraph{Baselines.} We compare our results with three versions of the stochastic rationalizer of \citet{lei2016rationalizing}: the original one -- \textbf{SFE} -- which uses the score function estimator to estimate the gradients; a second one -- \textbf{SFE w/ Baseline}  -- which uses SFE with a moving average baseline variance reduction technique; a third -- \textbf{Gumbel} -- in which we employ the Gumbel-Softmax reparameterization \citep{jang2017categorical} to reparameterize the Bernoulli variables; and, a fourth -- \textbf{HardKuma} -- in which we employ HardKuma variables \citep{bastings2019interpretable} instead of Bernoulli variables and use reparameterized gradients for training end-to-end. Moreover, the latter rationalizer employs a Lagrangian relaxation to solve the constrained optimization problem of targeting specific sparsity rates.  We also experimented with two deterministic strategies that use sparse attention mechanisms: a first that utilizes \textbf{sparsemax} \citep{martinssparsemax}, and a second that utilizes \textbf{fusedmax} \citep{niculae2019regularized} which encourages the network to pay attention to contiguous segments of text, by adding an additional total variation regularizer, inspired by the fused lasso. It is a natural deterministic counterpart of the constrained rationalizer proposed by \citet{lei2016rationalizing}, since the regularization encourages both sparsity and contiguity. The use of fusedmax for this task is new to the best of our knowledge. Similarly to \citet{jain2020learning}, we found that the stochastic rationalizers of \citet{lei2016rationalizing} and its variants (SFE, SFE w/ Baseline and Gumbel) require cumbersome hyperparameter search and tend to degenerate in such a way that the generated rationales are either the whole input text or empty text. Thus, at inference time, we follow the strategy proposed by \citet{jain2020learning} and restrict the generated rationale to a specified length $\ell$ via two mappings: \textbf{contiguous}, in which the span of length $\ell$, out of all the spans of this length, whose token-level scores cumulative sum is the highest is selected; and $\bm{\textbf{top-}k}$, in which the $\ell$ tokens with highest token-level scores are selected. Contrary to \citep{jain2020learning}, for the rationalizer of \citet{bastings2019interpretable} (HardKuma), we carefully tuned both the model hyperparameters and the Lagrangian relaxation algorithm hyperparameters, so as to use the deterministic policy in testing time that they propose.\footnote{We have found that using the deterministic policy at testing time proposed by \citet{bastings2019interpretable} instead of the top-$k$ or contiguous strategies is critical to achieve good performance with the HardKuma rationalizer.} All implementation details can be found in \S\ref{sec:implementation_details}. We also report the full-text baselines for each dataset in \S\ref{app:vanilla_classifiers}.

\subsection{Matchings for Natural Language Inference}
\paragraph{Data and Evaluation.} We used the English language SNLI and MNLI datasets \citep{bowman2015large, chen-etal-2017-enhanced}. We evaluate end task performance for both datasets. For the experiments with the {\textsf{M:Budget}}, we used a fixed budget of $B=4$ for SNLI and $B=6$ for MNLI. We also conduct further experiments with the HANS dataset \citep{mccoy2019right} which aims to analyse the use of linguistic heuristics (lexical overlap, constituent and subsequence heuristics) of NLI systems. The statistics and details of each dataset can be found in \S\ref{sec:data_matchings}. 

\paragraph{Baselines.} We compare our results with variants of constrained optimal transport for selective rationalization employed by \citet{swanson2020rationalizing}: relaxed 1:1, which is similar in nature to our proposed {\textsf{M:AtMostOne2}} factor; and exact $k=4$ similar to our proposed {\textsf{M:Budget}} with budget $B=4$. We also replicate the LP-matching implementation of \citet{niculae2020lpsparsemap} which consists of the original ESIM model described in \S \ref{sec:matchings-extraction} with $\bm{Z}$ as the output of the LP-SparseMAP problem with a {\textsf{M:XorAtMostOne}} factor. Importantly, both these models aggregate the encoded premise representation with the information that comes from the alignment. All implementation details can be found in \S\ref{sec:implementation_details}. We also report the ESIM baselines in \S\ref{app:vanilla_classifiers}.

\section{Results and Analysis}
\label{sec:results}
\subsection{Extraction of Text Highlights}
\renewcommand{\arraystretch}{.95}
\begin{table*}[t]
\footnotesize
\centering
\begin{tabular}{>{\arraybackslash}m{0.01cm}
>{\arraybackslash}m{2.35cm} >{\arraybackslash}m{1.5cm}
>{\raggedleft\arraybackslash}m{1.6cm}
>{\raggedleft\arraybackslash}m{2cm} 
>{\raggedleft\arraybackslash}m{1.7cm}
>{\raggedleft\arraybackslash}m{1.9cm} 
>{\raggedleft\arraybackslash}m{1.5cm}}
\toprule
Method &  & Rationale & SST $\uparrow$  & AgNews $\uparrow$ & IMDB $\uparrow$ & Beer $\downarrow$ & Hotels $\uparrow$\\ \midrule
\multirow{2}{*}{\stochastic} & \multirow{2}{*}{SFE} & top-$k$ & \multicolumn{1}{r}{.76 \textcolor{black!90}{\scriptsize{(.71/.80)}}} & \multicolumn{1}{r}{.92 \textcolor{black!90}{\scriptsize{(.92/.92)}}} & \multicolumn{1}{r}{.84 \textcolor{black!90}{\scriptsize{(.72/.88)}}} & \multicolumn{1}{r}{.018 \textcolor{black!90}{\scriptsize{(.016/.020)}}} & \multicolumn{1}{r}{.66 \textcolor{black!90}{\scriptsize{(.62/.69)}}}\\
& & contiguous & \multicolumn{1}{r}{.71 \textcolor{black!90}{\scriptsize{(.68/.75)}}} & \multicolumn{1}{r}{.86 \textcolor{black!90}{\scriptsize{(.85/.86)}}} & \multicolumn{1}{r}{.65 \textcolor{black!90}{\scriptsize{(.57/.73)}}} & \multicolumn{1}{r}{.020 \textcolor{black!90}{\scriptsize{(.019/.024)}}} & \multicolumn{1}{r}{.62 \textcolor{black!90}{\scriptsize{(.34/.72)}}}\\ \midrule
\multirow{2}{*}{\stochastic} & \multirow{2}{*}{SFE w/ Baseline} & top-$k$ & \multicolumn{1}{r}{.78 \textcolor{black!90}{\scriptsize{(.76/.80)}}} & \multicolumn{1}{r}{.92 \textcolor{black!90}{\scriptsize{(.92/.93)}}} & \multicolumn{1}{r}{.82 \textcolor{black!90}{\scriptsize{(.72/.88)}}} & \multicolumn{1}{r}{.019 \textcolor{black!90}{\scriptsize{(.017/.020)}}} & \multicolumn{1}{r}{.56 \textcolor{black!90}{\scriptsize{(.34/.64)}}}\\
& & contiguous & \multicolumn{1}{r}{.70 \textcolor{black!90}{\scriptsize{(.64/.75)}}} & \multicolumn{1}{r}{.86 \textcolor{black!90}{\scriptsize{(.84/.86)}}} & \multicolumn{1}{r}{.76 \textcolor{black!90}{\scriptsize{(.73/.80)}}} & \multicolumn{1}{r}{.021 \textcolor{black!90}{\scriptsize{(.019/.025)}}} & \multicolumn{1}{r}{.55 \textcolor{black!90}{\scriptsize{(.34/.69)}}}\\ \midrule
\multirow{2}{1.45cm}{\stochastic} & \multirow{2}{1.45cm}{Gumbel} & top-$k$ & \multicolumn{1}{r}{.70 (\textcolor{black!90}{\scriptsize{.67/.72)}}} & \multicolumn{1}{r}{.78 \textcolor{black!90}{\scriptsize{(.73/.84)}}} & \multicolumn{1}{r}{.74 \textcolor{black!90}{\scriptsize{(.71/.78)}}} & \multicolumn{1}{r}{.026 \textcolor{black!90}{\scriptsize{(.018/.041)}}} & \multicolumn{1}{r}{.83 \textcolor{black!90}{\scriptsize{(.73/.92)}}} \\
& & contiguous & \multicolumn{1}{r}{.67 \textcolor{black!90}{\scriptsize{(.67/.68)}}} & \multicolumn{1}{r}{.77 \textcolor{black!90}{\scriptsize{(.74/.81)}}} & \multicolumn{1}{r}{.72 \textcolor{black!90}{\scriptsize{(.72/.73)}}} & \multicolumn{1}{r}{.043 \textcolor{black!90}{\scriptsize{(.040/.048)}}} & \multicolumn{1}{r}{.74 \textcolor{black!90}{\scriptsize{(.65/.84)}}} \\ \midrule
\multirow{1}{1.45cm}{\stochastic} & \multirow{1}{1.45cm}{HardKuma} & -- & \multicolumn{1}{r}{.80 \textcolor{black!90}{\scriptsize{(.80/.81)}}} & \multicolumn{1}{r}{.90 \textcolor{black!90}{\scriptsize{(.87/.88)}}} & \multicolumn{1}{r}{.87 \textcolor{black!90}{\scriptsize{(.90/.91)}}} & \multicolumn{1}{r}{.019 \textcolor{black!90}{\scriptsize{(.016/.020)}}} & \multicolumn{1}{r}{.90 \textcolor{black!90}{\scriptsize{(.88/.92)}}} \\ \midrule
\multirow{2}{1.45cm}{\deterministic} & \multirow{2}{2.85cm}{Sparse Attention} & sparsemax & \multicolumn{1}{r}{\textbf{.82} \textcolor{black!90}{\scriptsize{(.81/.83)}}} & \multicolumn{1}{r}{\textbf{.93} \textcolor{black!90}{\scriptsize{(.93/.93)}}} & \multicolumn{1}{r}{.89 \textcolor{black!90}{\scriptsize{(.89/.90)}}} & \multicolumn{1}{r}{.019 \textcolor{black!90}{\scriptsize{(.016/.021)}}} & \multicolumn{1}{r}{.89 \textcolor{black!90}{\scriptsize{(.87/.92)}}} \\ 
& & fusedmax & \multicolumn{1}{r}{.81 \textcolor{black!90}{\scriptsize{(.81/.82)}}} & \multicolumn{1}{r}{.92 \textcolor{black!90}{\scriptsize{(.91/.92)}}} & \multicolumn{1}{r}{.88 \textcolor{black!90}{\scriptsize{(.87/.89)}}} & \multicolumn{1}{r}{{.018} \textcolor{black!90}{\scriptsize{(.017/.019)}}} & \multicolumn{1}{r}{.85 \textcolor{black!90}{\scriptsize{(.77/.90)}}}\\ \midrule
\deterministic & SPECTRA (ours) & H:SeqBudget &  \multicolumn{1}{r}{.80 \textcolor{black!90}{\scriptsize{(.79/.81)}}} & \multicolumn{1}{r}{.92 \textcolor{black!90}{\scriptsize{(.92/.93)}}} & \multicolumn{1}{r}{\textbf{.90} \textcolor{black!90}{\scriptsize{(.89/.90)}}} & \multicolumn{1}{r}{\textbf{.017} \textcolor{black!90}{\scriptsize{(.016/.019)}}} & \multicolumn{1}{r}{\textbf{.91} \textcolor{black!90}{\scriptsize{(.90/.92)}}}\\ \bottomrule
\end{tabular}
\caption{\label{tab:predictive_performance}
Model predictive performances across datasets, for stochastic (\stochastic) and deterministic (\deterministic) methods. We report mean and min/max $F_1$ scores across five random seeds on test sets for all datasets but Beer where we report MSE.  We bold the best-performing rationalized model(s) for each corpus.\vspace{-6pt}
}
\end{table*}
\renewcommand{\arraystretch}{.8}
\begin{table*}[t]
\footnotesize
\centering
\begin{tabular}{>{\arraybackslash}m{0.01cm}
>{\arraybackslash}m{2.35cm} >{\arraybackslash}m{1.5cm}
>{\raggedleft\arraybackslash}m{1.6cm}
>{\raggedleft\arraybackslash}m{2cm} 
>{\raggedleft\arraybackslash}m{1.7cm}
>{\raggedleft\arraybackslash}m{1.9cm} 
>{\raggedleft\arraybackslash}m{1.5cm}}
\toprule
 Method & & Rationale & SST & AgNews & IMDB & Beer & Hotels \\ \midrule
 \stochastic & HardKuma & -- &  \multicolumn{1}{r}{.15 \textcolor{black!90}{\scriptsize{(.12/.19)}}} & \multicolumn{1}{r}{.19 \textcolor{black!90}{\scriptsize{(.18/.19)}}} & \multicolumn{1}{r}{{.03} \textcolor{black!90}{\scriptsize{(.02/.03)}}} & \multicolumn{1}{r}{{.08} \textcolor{black!90}{\scriptsize{(.00/.17)}}} & \multicolumn{1}{r}{{.09} \textcolor{black!90}{\scriptsize{(.07/.12)}}}\\ \midrule
\multirow{2}{*}{\deterministic} &\multirow{2}{2.85cm}{Sparse Attention} & sparsemax & \multicolumn{1}{r}{.17 \textcolor{black!90}{\scriptsize{(.13/.23)}}} & \multicolumn{1}{r}{.13 \textcolor{black!90}{\scriptsize{(.11/.15)}}} & \multicolumn{1}{r}{.02 \textcolor{black!90}{\scriptsize{(.02/.03)}}} & \multicolumn{1}{r}{.11 \textcolor{black!90}{\scriptsize{(.09/.13)}}} & \multicolumn{1}{r}{.03 \textcolor{black!90}{\scriptsize{(.02/.04)}}} \\ 
& & fusedmax & \multicolumn{1}{r}{.60 \textcolor{black!90}{\scriptsize{(.14/1.0)}}} & \multicolumn{1}{r}{.32 \textcolor{black!90}{\scriptsize{(.10/.66)}}} & \multicolumn{1}{r}{.02 \textcolor{black!90}{\scriptsize{(.01/.02)}}} & \multicolumn{1}{r}{{.26} \textcolor{black!90}{\scriptsize{(.03/.98)}}} & \multicolumn{1}{r}{.04 \textcolor{black!90}{\scriptsize{(.01/.08)}}}\\ \bottomrule

\end{tabular}
\caption{\label{tab:avg_size_rationale}
Average size of the extracted rationales using the HardKuma stochastic rationalizer (\stochastic) and deterministic (\deterministic) sparse attention mechanisms. We report mean and min/max average size across five random seeds.\vspace{-6pt}
}
\end{table*}
\paragraph{Predictive Performance.} We report the predictive performances of all models in Table \ref{tab:predictive_performance}. We observe that the deterministic rationalizers that use sparse attention mechanisms generally outperform the stochastic rationalizers while exhibiting lower variability across different random seeds and different datasets. In general and as expected, for the stochastic models, the top-$k$ strategy for rationale extraction outperforms the contiguous strategy. As reported in \citet{jain2020learning}, strategies that impose a contiguous mapping trade coherence for performance on the end-task. Our experiments also show that HardKuma is the stochastic rationalizer least prone to variability across different seeds, faring competitively with the deterministic methods. The strategy proposed in this paper, \textsf{{H:SeqBudget}}, fares competitively with the deterministic methods and generally outperforms the stochastic methods. Moreover, similarly to the other deterministic rationalizers, our method exhibits lower variability across different runs. 
We show examples of highlights extracted by SPECTRA in \S\ref{app:example_highlights_rationales}.

\subsection{Quality of the Rationales} 

\paragraph{Rationale Regularization.} We report in Table \ref{tab:avg_size_rationale} the average size of the extracted rationales (proportion of words
not zeroed out) across datasets for the stochastic HardKuma rationalizer and for each rationalizer that uses sparse attention mechanisms. The latter strategies do not have any mechanism to regularize the sparsity of the extracted rationales, which leads to variability on the rationale extraction. This is especially the case for the fusedmax strategy, as it pushes adjacent tokens to be given the same attention probability. This might lead to rationale degeneration when the attention weights are similar across all tokens. On the other hand, HardKuma employs a Lagrangian relaxation algorithm to target a predefined sparsity level. We have found that careful hyperparameter tuning is required across different datasets. While, generally, the average size of the extracted rationales does not exhibit considerable variability,  some random seeds led to degeneration (the model extracts empty rationales). Remarkably, our proposed strategy utilizes the $\mathsf{BUDGET}$ factor to set a predefined desired rationale length, regularizing the rationale extraction while still applying a deterministic policy that exhibits low variability across different runs and datasets (Table \ref{tab:predictive_performance}). 

\paragraph{Matching with Human Annotations.} We report token-level $F_1$ scores  in Table \ref{tab:rationale_matching} to evaluate the quality of the rationales for the datasets for which we had human annotations for the test set. We observe that our proposed strategy and HardKuma outperform all the other methods on what concerns matching the human annotations. This was to be expected considering the results shown in Table \ref{tab:predictive_performance} and Table \ref{tab:avg_size_rationale}: 
the stochastic models other than HardKuma do not fare competitively with the deterministic models and their variability across runs is also reflected on the token-level $F_1$ scores; and although the rationalizers that use sparse attention mechanisms are competitive with our proposed strategy, the lack of regularization on what comes to the rationale extraction leads to variable sized rationales which is also reflected on poorer matchings. We also observe that, when degeneration does not occur, HardKuma generally extracts high quality rationales on what comes to matching the human annotations. It is also worth remarking that the sparsemax and top-$k$ strategies are not expected to fare well on this metric because human annotations for these datasets are at the \textit{sentence-level}. Our strategy, however, not only pushes for sparser rationales but also encourages contiguity on the extraction.
\renewcommand{\arraystretch}{.9}
\begin{table}[t]
\footnotesize
\centering
\begin{tabular}{>{\arraybackslash}m{0.01cm} 
>{\arraybackslash}m{1.45cm} >{\arraybackslash}m{1.45cm}
>{\raggedleft\arraybackslash}m{1.2cm}
>{\raggedleft\arraybackslash}m{1.2cm}}
\toprule
 Method & & Rationale & Beer & Hotels \\ \midrule
\multirow{2}{*}{\stochastic} & \multirow{2}{*}{SFE} & top-$k$ & \multicolumn{1}{r}{.19 \textcolor{black!90}{\scriptsize{(.13/.30)}}} & \multicolumn{1}{r}{.16 \textcolor{black!90}{\scriptsize{(.12/.30)}}} \\
& & contiguous & \multicolumn{1}{r}{.35 \textcolor{black!90}{\scriptsize{(.18/.42)}}} & \multicolumn{1}{r}{.14 \textcolor{black!90}{\scriptsize{(.12/.15)}}} \\ \midrule
\multirow{2}{*}{\stochastic} & \multirow{2}{1.45cm}{SFE w/ Baseline} & top-$k$ & \multicolumn{1}{r}{.17 \textcolor{black!90}{\scriptsize{(.14/.19)}}} & \multicolumn{1}{r}{.14 \textcolor{black!90}{\scriptsize{(.13/.18)}}} \\
& & contiguous & \multicolumn{1}{r}{.41 \textcolor{black!90}{\scriptsize{(.37/.42)}}} & \multicolumn{1}{r}{.15 \textcolor{black!90}{\scriptsize{(.14/.15)}}} \\ \midrule
\multirow{2}{1.85cm}{\stochastic} & \multirow{2}{1.85cm}{Gumbel} & top-$k$ & \multicolumn{1}{r}{.27 \textcolor{black!90}{\scriptsize{(.14/.39)}}} & \multicolumn{1}{r}{.36 \textcolor{black!90}{\scriptsize{(.27/.48)}}} \\
& & contiguous & \multicolumn{1}{r}{.42 \textcolor{black!90}{\scriptsize{(.41/.42)}}} & \multicolumn{1}{r}{.36 \textcolor{black!90}{\scriptsize{(.29/.48)}}} \\ \midrule
\multirow{1}{1.85cm}{\stochastic} & \multirow{1}{1.85cm}{HardKuma} & -- & \multicolumn{1}{r}{.37 \textcolor{black!90}{\scriptsize{(.00/.90)}}}    & \multicolumn{1}{r}{\textbf{.52} \textcolor{black!90}{\scriptsize{(.37/.57)}}} \\ \midrule
\multirow{2}{1.85cm}{\deterministic} & \multirow{2}{1.45cm}{Sparse Attention} & sparsemax  & \multicolumn{1}{r}{.48 \textcolor{black!90}{\scriptsize{(.41/.55)}}} & \multicolumn{1}{r}{.17 \textcolor{black!90}{\scriptsize{(.07/.31)}}} \\ 
&  & fusedmax & \multicolumn{1}{r}{.39 \textcolor{black!90}{\scriptsize{(.29/.53)}}}  & \multicolumn{1}{r}{.25 \textcolor{black!90}{\scriptsize{(.09/.31)}}} \\ \midrule
\deterministic & SPECTRA \newline (ours) & H:SeqBudget &  \multicolumn{1}{r}{\textbf{.61} \textcolor{black!90}{\scriptsize{(.56/.68)}}} & \multicolumn{1}{r}{.37 \textcolor{black!90}{\scriptsize{(.34/.40)}}}\\ \bottomrule
\end{tabular}
\caption{
Evaluation of the rationales through matching with human annotations, for stochastic (\stochastic) and deterministic (\deterministic) methods. We report mean token-level $F_1$ scores and min/max across five random seeds.\vspace{-6pt}
}
\label{tab:rationale_matching}
\end{table}
\subsection{Extraction of Text Matchings}

\renewcommand{\arraystretch}{.9}
\begin{table}[t]
\footnotesize
\centering
\begin{tabular}{
>{\arraybackslash}m{3.3cm}
>{\raggedleft\arraybackslash}m{1.1cm}
>{\raggedleft\arraybackslash}m{1.1cm} }
\toprule
Matching Structure & SNLI & MNLI \\ \midrule
\textit{\textbf{Not Faithful}} & & \\
\deterministic~OT relaxed 1:1$^\dagger$ & \multicolumn{1}{r}{.82} & \multicolumn{1}{r}{--} \\ 
\deterministic~OT exact $k=4^\dagger$ & \multicolumn{1}{r}{.81} & \multicolumn{1}{r}{--} \\\midrule
\stochastic~Gumbel Matching & \multicolumn{1}{r}{.85 \textcolor{black!90}{\scriptsize{(.84/.85)}}}  & \multicolumn{1}{r}{.73 \textcolor{black!90}{\scriptsize{(.72/.73)}}}  \\ 
\deterministic~M:XorAtMostOne &  \multicolumn{1}{r}{\textbf{.86} \textcolor{black!90}{\scriptsize{(.86/.87)}}} & \multicolumn{1}{r}{\textbf{.76} \textcolor{black!90}{\scriptsize{(.75/.76)}}}\\ 
\deterministic~M:AtMostOne2 &  \multicolumn{1}{r}{\textbf{.86} \textcolor{black!90}{\scriptsize{(.86/.87)}}} & \multicolumn{1}{r}{\textbf{.76} \textcolor{black!90}{\scriptsize{(.75/.76)}}} \\
\deterministic~M:Budget &  \multicolumn{1}{r}{.85 \textcolor{black!90}{\scriptsize{(.85/.86)}}} & \multicolumn{1}{r}{.75 \textcolor{black!90}{\scriptsize{(.75/.76)}}} \\ \midrule\midrule
\textit{\textbf{Faithful}} & & \\
\stochastic~Gumbel Matching & \multicolumn{1}{r}{.85 \textcolor{black!90}{\scriptsize{(.84/.85)}}} & \multicolumn{1}{r}{.73 \textcolor{black!90}{\scriptsize{(.72/.73)}}} \\ 
\deterministic~M:XorAtMostOne & \multicolumn{1}{r}{.85 \textcolor{black!90}{\scriptsize{(.85/.85)}}}  & \multicolumn{1}{r}{.73 \textcolor{black!90}{\scriptsize{(.72/.73)}}} \\
\deterministic~M:AtMostOne2 &  \multicolumn{1}{r}{.85 \textcolor{black!90}{\scriptsize{(.85/.85)}}} & \multicolumn{1}{r}{.73 \textcolor{black!90}{\scriptsize{(.73/.73)}}}\\
\deterministic~M:Budget &  \multicolumn{1}{r}{ .82 \textcolor{black!90}{\scriptsize{(.81/.82)}}} & \multicolumn{1}{r}{ .68 \textcolor{black!90}{\scriptsize{(.67/.68)}}}\\\bottomrule

\end{tabular}
\caption{
Model predictive performances across datasets, for stochastic (\stochastic) and deterministic (\deterministic) methods.  We report mean and min/max $F_1$ scores across three random seeds on both SNLI and MNLI test sets.  We bold the best-performing rationalized models for each corpus. $\dagger$ means results come from \citet{swanson2020rationalizing}.
}
\label{tab:results_predperformance_matchings}
\end{table}

\renewcommand{\arraystretch}{.7}
\begin{table}[h!]
\footnotesize
\centering
\begin{tabular}{
>{\arraybackslash}m{2.9cm}
>{\raggedleft\arraybackslash}m{1.75cm}
>{\raggedleft\arraybackslash}m{1.75cm} }
\toprule
HANS Subcomponent & Vanilla & Augmented \\ \midrule
\textit{\textbf{Entailment}} & & \\
Lexical Overlap & \multicolumn{1}{r}{.9942} & \multicolumn{1}{r}{.9962} \\ 
Subsequence & \multicolumn{1}{r}{.9960} & \multicolumn{1}{r}{1.0} \\
Constituent & \multicolumn{1}{r}{.9988} & \multicolumn{1}{r}{1.0} \\\midrule \midrule
\textit{\textbf{Non-entailment}} & & \\
Lexical Overlap & \multicolumn{1}{r}{.0052} & \multicolumn{1}{r}{.9998} \\ 
Subsequence & \multicolumn{1}{r}{.0016} & \multicolumn{1}{r}{1.0} \\
Constituent & \multicolumn{1}{r}{.0122} & \multicolumn{1}{r}{1.0} \\\bottomrule

\end{tabular}
\caption{
Model predictive performances for the vanilla and augmented models evaluated on the HANS evaluation set.  We report accuracies for each of the six subcomponents of the evaluation set.\vspace{-5pt}
}
\label{tab:results_HANS}
\end{table}

\paragraph{Predictive Performance.} We report the predictive performances of all models in Table \ref{tab:results_predperformance_matchings}. Both the strategies that use the LP-SparseMAP extraction layer and our proposed stochastic matchings extractor outperform the OT variants for matchings extraction.  We observe that, contrary to the text highlights experiments, the stochastic matchings extraction model does not exhibit noticeably higher variability compared to the deterministic models. In general, the faithful models are competitive with the non-faithful models. Since the latter ones are constrained to only utilize information from the premise that comes from alignments, these results demonstrate the effectiveness of the alignment extraction. As expected, there is a slight trade-off between how constrained the alignment is and the model's predictive performance. This is more noticeable with the M:Budget strategy, the most constrained version of our proposed strategies, in the faithful scenario. We show examples of matchings extracted by SPECTRA in \S\ref{app:example_matchings_rationales}.

\paragraph{Heuristics Analysis with HANS.} We used two different {\textsf{M:AtMostOne2}} models for our analysis: a first one trained on MNLI (\textbf{Vanilla}), and a second one trained on MNLI augmented (\textbf{Augmented}) with 30,000 HANS-like examples ($\approx$ 8\% of MNLI original size), replicating the data augmentation scenario in \cite{mccoy2019right}. We evaluated both models on the HANS evaluation set, which has six subcomponents, each defined by its correct label and the heuristic it addresses. We report the results in Table \ref{tab:results_HANS}. Our models behave similarly to those in \cite{mccoy2019right}: when we augment the training set with HANS-like examples, the model no longer associates the heuristics to entailment. By observation of the extracted matchings, we noticed that these were similar between the two models. Thus, the effect of the augmented data resides on how the information from the matchings is used after the extraction layer. We show examples of matchings in \S\ref{app:example_matchings_rationales}.

\section{Related Work}
\label{sec:related}

\paragraph{Selective Rationalization.} There is a long string of work on interpreting predictions made by neural networks \citep{lipton2017mythos, doshivelez2017rigorous, gilpin2019explaining, wiegreffe2021teach, zhang2021survey}.  
Our paper focus on selective rationalizers, which have been used for extraction of text highlights \citep{lei2016rationalizing, bastings2019interpretable, yu2019rethinking, deyoung2019eraser, treviso-martins-2020-explanation, Zhang_2021} and text matchings \citep{swanson2020rationalizing}. Most works rely 
on stochastic rationale generation or deterministic attention mechanisms, 
but the two approaches have never been extensively compared. 
Our work adds that comparison and contributes with an easy-to-train fully differentiable rationalizer that allows for flexible constrained rationale extraction. 
Our strategy for rationalization based on sparse structured prediction on factor graphs constitutes a unified framework for deterministic extraction of different structured rationales. 

\paragraph{Structured Prediction on Factor Graphs.} \citet{kim2017structured} incorporate structured models in attention mechanisms as a way to model rich structural dependencies, leading to a dense probability distribution over structures. \citet{niculae2018sparsemap} propose SparseMAP, which yields a sparse probability distribution over structures and can be computed using calls to a MAP oracle, making it applicable to problems (e.g. matchings) for which marginal inference is intractable but MAP is not. 
However, the requirement of an exact MAP oracle prohibits its application for more expressive structured models such as loopy graphical models and logic constraints. 
This limitation is overcome by LP-SparseMAP \citep{niculae2020lpsparsemap}
via a local polytope relaxation, extending the previous method to sparse differentiable optimization in any factor graph with arbitrarily complex structure. While other relaxations for matchings -- such as entropic regularization leading to Sinkhorn's algorithm \citep{cuturi-sinkhorn} -- that are tractable and efficient exist and have been used for rationalization \citep{swanson2020rationalizing}, we use LP-SparseMAP for rationale extraction in our work. Our approach for rationalization focuses on learning and explaining with latent structure extracted by structured prediction on factor graphs.

\paragraph{Sentence Compression and Summarization.} Work on sentence compression and summarization bears some resemblance to selective rationalization for text highlights extraction. \citet{titov-mcdonald-2008-joint} propose a statistical model which is able to discover corresponding topics in text and extract informative snippets of text by predicting a stochastic mask via Gibbs sampling. \citet{mcdonald-2006-discriminative} proposes a budgeted dynamic program in the same vein as that of the H:SeqBudget strategy for text highlights extraction. \citet{berg-kirkpatrick-etal-2011-jointly} and \citet{almeida-martins-2013-fast} propose models that jointly extract and compress sentences. Our work differs in that our setting is completely unsupervised and we need to differentiate through the extractive layers.


\section{Conclusions}
\label{sec:conclusions}

We have proposed SPECTRA, an easy-to-train fully differentiable rationalizer that allows for flexible constrained rationale extraction.
We have provided a comparative study with stochastic and deterministic approaches for rationalization, showing that SPECTRA generally outperforms previous rationalizers in text classification and natural language inference tasks. Moreover, it does so while exhibiting less variability than stochastic methods and easing regularization of the rationale extraction when compared to previous deterministic approaches. 
Our framework constitutes a unified framework for deterministic extraction of different structured rationales. We hope that our work spurs future research on rationalization for different structured explanations.

\section*{Acknowledgements}
We are grateful to Vlad Niculae for his valuable help and insight on LP-SparseMAP. We would like to thank Marcos Treviso for helping to start this project. We are grateful to Wilker Aziz, António Farinhas, Ben Peters, Gonçalo Correia, and the reviewers, for their helpful feedback and discussions. This work was supported by the European Research Council (ERC StG DeepSPIN 758969, by the FCT
through contract UIDB/50008/2020, and by the P2020 programs MAIA and Unbabel4EU (LISBOA-01-0247-FEDER-045909 and LISBOA-01-0247-FEDER-042671).

\newpage

\bibliography{corrected_anthology}
\bibliographystyle{acl_natbib}

\vfill
~
\pagebreak

\appendix
\newpage
\section{Datasets for Highlights Extraction}
\label{sec:data_highlights}

We used five datasets for sentiment analysis: four for text classification (SST, AgNews, IMDB, Hotels) \citep{socher-etal-2013-recursive, delcorso-agnews, maas-EtAl:2011:ACL-HLT2011, hotel-wang} and one for regression (BeerAdvocate) \cite{mcauley2012learning}.  The Hotels and BeerAdvocate datasets contain data instances for multiple aspects. In this work, we use the Hotels' location aspect and the BeerAdvocate's appearance aspect. These two datasets contain \textit{sentence-level} rationale annotations for their test sets. For these datasets, we use the splits used in \citet{bao2018deriving}. For all other datasets, we use the splits in \citet{2020HuggingFace-datasets}. For IMDB and AgNews we randomly selected 10\%, 15\% of examples from the training set to be used as validation data, respectively. Table \ref{tab:highlights_datasets} shows statistics for each dataset and rationale length as a percentage of each document.

\renewcommand{\arraystretch}{1}
\begin{table}[h!]
\footnotesize
\centering
\begin{tabular}{
>{\arraybackslash}m{.7cm}
>{\raggedleft\arraybackslash}m{1.05cm}
>{\raggedleft\arraybackslash}m{1.05cm} 
>{\raggedleft\arraybackslash}m{1.20cm}
>{\raggedleft\arraybackslash}m{1.45cm}}
\toprule
Dataset & \# Train & \# Test & \# Classes & Rationale Length (\%)\\ \midrule
SST &  \multicolumn{1}{r}{6920} & \multicolumn{1}{r}{1821} & \multicolumn{1}{r}{2} & 20 \\
AgNews & \multicolumn{1}{r}{120K} & \multicolumn{1}{r}{7600} & \multicolumn{1}{r}{4} & 20 \\
IMDB & \multicolumn{1}{r}{25K} & \multicolumn{1}{r}{25K} & \multicolumn{1}{r}{2} & 20\\
Hotels & \multicolumn{1}{r}{12K} & \multicolumn{1}{r}{200} & \multicolumn{1}{r}{2} & 15 \\
Beer & \multicolumn{1}{r}{80K} & \multicolumn{1}{r}{997} & -- & 10 \\
\bottomrule

\end{tabular}
\caption{\label{tab:highlights_datasets}
Dataset statistics and rationale length as a percentage of each document.
}
\end{table}

For the datasets without human annotations, we used the same sparsity level (20\%) -- \citet{jain2020learning} uses this value for AgNews and SST; for BeerAdvocate, we used the sparsity levels used in \citet{lei2016rationalizing} and \citet{yu2019rethinking}; and, for Hotels we opted to select a sparsity level of 15\% (human annotations average around 10\% sparsity level).

\section{Datasets for Matchings Extraction}
\label{sec:data_matchings}
For natural language inference (NLI), we used SNLI and MNLI \citep{bowman2015large, chen-etal-2017-enhanced}. For MNLI, we split the MNLI matched validation set into equal validation and test sets. Table \ref{tab:matchings_datasets} shows statistics for each dataset and the alignment budget used for the \textsf{{M:Budget}} factor.

\renewcommand{\arraystretch}{1}
\begin{table}[h!]
\footnotesize
\centering
\begin{tabular}{
>{\arraybackslash}m{.7cm}
>{\raggedleft\arraybackslash}m{1.05cm}
>{\raggedleft\arraybackslash}m{1.05cm} 
>{\raggedleft\arraybackslash}m{1.20cm}
>{\raggedleft\arraybackslash}m{1.45cm}}
\toprule
Dataset & \# Train & \# Test & \# Classes & Alignment Budget\\ \midrule
SNLI &  \multicolumn{1}{r}{550K} & \multicolumn{1}{r}{10K} & \multicolumn{1}{r}{3} & 4 \\
MNLI & \multicolumn{1}{r}{392K} & \multicolumn{1}{r}{10K} & \multicolumn{1}{r}{3} & 6 \\
HANS & \multicolumn{1}{r}{30K} & \multicolumn{1}{r}{30K} & \multicolumn{1}{r}{2} & -- \\
\bottomrule

\end{tabular}
\caption{\label{tab:matchings_datasets}
Dataset statistics and alignment budget for each dataset.
}
\end{table}

For SNLI, we set the Budget $B$ to 4 to compare with the OT approach (OT exact $k=4$) of \citet{swanson2020rationalizing}. For MNLI, we set B to 6, since the average premise length in MNLI is around 50\% bigger than that of SNLI.

We also conduct experiments with the HANS \citep{mccoy2019right} dataset. This dataset consists of a controlled evaluation set to detect whether NLI systems are exploring linguistic heuristics such as lexical overlap, subsequence and constituent heuristics. A detailed description of each of these heuristics can be found in the original paper. The dataset is also constituted by 30,000 HANS-like examples that can be used to augment existing NLI training sets such as SNLI or MNLI.

\section{Implementation Details}
\label{sec:implementation_details}


\subsection{Rationalizers Experimental Setup}
For all rationalizers, we map each input word to 300D-pretrained GloVe embeddings from 840B release \citep{pennington2014glove} that are kept frozen. We instantiate all encoder networks as bidirectional LSTM \citep{LSTM} layers (BiLSTM) (w/ hidden size 200) similarly to \citet{lei2016rationalizing, bastings2019interpretable, treviso-martins-2020-explanation}. Although other works \citep{jain2020learning, paranjape-etal-2020-information} use more powerful BERT-based representations, we firstly experimented with BiLSTM layers and noticed our results were competitive with those reported in \citet{jain2020learning}. We used the Adam optimizer \citep{kingma2017adam} for all experiments with learning rate within $\{1 \times 10^{-3}, 5 \times 10^{-4}, 1 \times 10^{-4}, 5 \times 10^{-5}\}$ and $\ell_2$ regularization within $\{10^{-4}, 10^{-5}\}$. We also enforce a grad norm of 5.0. We train all models for highlights extraction for a minimum of 5 epochs and maximum of 25 epochs. For matchings extraction, we set the number of minimum epochs and maximum epochs to 3 and 10, respectively.

Training for all methods for highlights extraction but HardKuma is stopped if Macro $F_1$ (for classification) or MSE (for regression) is not improved for 5 epochs. For matchings extraction, training is stopped if Macro $F_1$ does not improve for 3 epochs. For HardKuma, we train until the maximum number of epochs. This is because the rationale length might vary considerably during training due to the Lagrangian relaxation algorithm that is employed at training time. We found that using early stopping would often favour models that selected almost all of the input text. Unlike \citet{jain2020learning}, we decided to carefully tune both model and the Lagrangian relaxation algorithm hyperparameters for this rationalizer. This had a big impact on the performance, as HardKuma performed poorly with the top-$k$ and contiguous strategies at inference time. Even though some careful tuning is required and degeneration might occur for some random seeds, it is still much less cumbersome than tuning the variants of the rationalizer of \citet{lei2016rationalizing}. We hypothesize that this is mostly due to two factors: the control on the rationale average size that the Lagrangian relaxation algorithm aims to impose; and the gradient estimates with reparameterized gradients exhibit less variance than those with the score function estimator. 

All models for highlights extraction have 1.8M trainable parameters. Models for faithful and non-faithful selective rationalization of text matchings have 1.7M and 1.8M trainable parameters, respectively.

\subsection{SPECTRA Sparsity Regularization}
During training, we apply a temperature term $T$ in the \textsf{sparsemax} and \textsf{fusedmax} operators. This parameter is set within $\{0.05, 0.1, 0.2\}$. The total variation regularization for fusedmax is set to $0.7$. 

For the models that use the LP-SparseMAP extraction layer, we use a temperature term $T$ set within $\{0.05, 0.1, 0.2\}$ during training. Moreover, for the H:SeqBudget, we set the transition scores within $\{0.001, 0.005\}$ for all datasets. All hyperparamter searches were conducted manually.

The LP-SparseMAP problem can be interpreted as the $\ell_2$-regularized LP-MAP. Its output corresponds to a probability distribution over a sparse set of structures. Therefore LP-MAP can be seen as LP-SparseMAP with the scores divided by a zero-limit temperature parameter. This procedure at test time would lead to the LP-MAP solution, which is generally an outer relaxation of MAP \citep{martins2015ad3}. When inference in the factor graph is exact, the solutions of the LP-MAP are integer (i.e., LP-MAP yields the true MAP). But that is not the case for when inference in the factor graph is not exact. Thus, LP-SparseMAP solutions for this test time setting might be a soft or discrete selection of parts of the input. We used a temperature parameter of $10^{-3}$ at validation and testing time.

\subsection{Computing Infrastructure}
Our infrastructure consists of 2 machines with the specifications shown in Table \ref{tab:infrastructure}. The machines were used interchangeably, and all experiments were executed in a single GPU. We did not observe significant differences in the execution time of our models across different machines. 
\renewcommand{\arraystretch}{1}
\begin{table}[h]
\footnotesize
\centering
\begin{tabular}{
>{\arraybackslash}m{.25cm}
>{\raggedright\arraybackslash}m{3.05cm}
>{\raggedright\arraybackslash}m{3.05cm}}
\toprule
\# & GPU & CPU\\ \midrule
1 & 4×GTX 1080 Ti - 12GB & 8×Intel i7-9800X @ 3.80GHz - 128GB \\
2 & 4×RTX 2080 Ti - 12GB & 48xIntel Xeon Silver @ 2.20GHz - 128GB \\
\bottomrule 

\end{tabular}
\caption{\label{tab:infrastructure}
Computing infrastructure used to run our experiments.
}
\end{table}

\section{Full-text Baselines}
\label{app:vanilla_classifiers}

Table \ref{tab:full_text_baselines} shows the performance scores of each classifier on the test sets for highlights extraction when feeding the full input document to the predictor.

\renewcommand{\arraystretch}{1}
\begin{table}[h!]
\footnotesize
\centering
\begin{tabular}{
>{\arraybackslash}m{0.2\columnwidth} 
>{\raggedleft\arraybackslash}m{0.35\columnwidth}}
\toprule
 Dataset & Performance Score \\ \midrule
 SST  & \multicolumn{1}{r}{0.83 \textcolor{black!90}{\scriptsize{(.82/.83)}}}\\
 AgNews  & \multicolumn{1}{r}{0.93 \textcolor{black!90}{\scriptsize{(.93/.93)}}}
\\
IMDB & \multicolumn{1}{r}{0.90 \textcolor{black!90}{\scriptsize{(.90/.90)}}}\\
Beer  & \multicolumn{1}{r}{0.019 \textcolor{black!90}{\scriptsize{(.18/.21)}}} \\
Hotels &  \multicolumn{1}{r}{0.87 \textcolor{black!90}{\scriptsize{(.86/.88)}}}\\
\bottomrule

\end{tabular}
\caption{\label{tab:full_text_baselines}
Model predictive performances across datasets using full-text. We report mean and min/max $F_1$ scores across five random seeds on test sets for all datasets but Beer where we report MSE.}
\end{table}

Table \ref{tab:ESIM_baselines} shows the performance scores of the ESIM model for both SNLI and MNLI.

\renewcommand{\arraystretch}{1}
\begin{table}[h!]
\footnotesize
\centering
\begin{tabular}{
>{\arraybackslash}m{0.2\columnwidth} 
>{\raggedleft\arraybackslash}m{0.35\columnwidth}}
\toprule
 Dataset & Performance Score \\ \midrule
 SNLI  & \multicolumn{1}{r}{0.86 \textcolor{black!90}{\scriptsize{(.86/.86)}}}\\
 MNLI  & \multicolumn{1}{r}{0.74 \textcolor{black!90}{\scriptsize{(.73/.74)}}}
\\
\bottomrule

\end{tabular}
\caption{\label{tab:ESIM_baselines}
ESIM predictive performances for SNLI and MNLI. We report mean and min/max $F_1$ scores across three random seeds on test sets for both datasets.}
\end{table}

\section{Computational Cost of SPECTRA}
\label{app:comp_cost_spectra}
Tables \ref{tab:comp_times_highlights} and \ref{tab:comp_times_matchings} show the average training and validation time per epoch for each dataset with the SPECTRA strategies for highlights and matchings extraction, respectively. We also present, for comparison, the average training and validation time per epoch for some of the other methods we used in the paper.  The batch size for the models for highlights extraction is $32$. For matchings extraction, the batch size for all models but M:Budget is $16$. For M:Budget, the batch size is $8$.

The computational time of SPECTRA depends on several factors inherited from the use of LP-SparseMAP as the extractive method. Generally, the bigger the number of local factors $f \in \mathcal{F}$, the more costly it is to compute a solution. Thus, it might be necessary to increase the number of iterations for the LP-SparseMAP to converge to a solution for which all factors agree. We set this number to 10 in training time following \citet{niculae2020lpsparsemap}. During inference, we set a maximum number of iterations of 1000. For highlights extraction, the \textsf{H:SeqBudget} consists of a single factor, thus the solution is found within a single iteration. For matchings extraction, our factors consist of multiple local factors that impose hard constraints that must agree in the final matching: \textsf{M:XorAtMostOne} and \textsf{M:AtMostOne2} consist of $L_P + L_H$ local factors, and \textsf{M:Budget} adds an additional global budget factor to the factor graph of \textsf{M:AtMostOne2}, yielding a more complex overall problem. Faster times would be achieved for smaller values of maximum number of iterations.

\renewcommand{\arraystretch}{1}
\begin{table}[h!]
\footnotesize
\centering
\begin{tabular}{
>{\arraybackslash}m{1.25cm}
>{\raggedleft\arraybackslash}m{0.55cm}
>{\raggedleft\arraybackslash}m{1.0cm} 
>{\raggedleft\arraybackslash}m{0.75cm}
>{\raggedleft\arraybackslash}m{0.75cm}
>{\raggedleft\arraybackslash}m{0.75cm}}
\toprule
Strategy & SST & AgNews & IMDB & Beer & Hotels\\ \midrule
\multicolumn{6}{l}{\textbf{\textit{Training Time}} (in seconds)} \\ \midrule
SPECTRA &  20 & 300 & 600 & 85 & 550\\
HardKuma & 18 & 180 & 120 & 75 & 220 \\
SFE &  10 & 165 & 120 & 40 & 200\\
Sparsemax & 15 & 165 & 120 &  40 & 200\\
\midrule \midrule
\multicolumn{5}{l}{\textbf{\textit{Validation Time}} (in seconds)} \\ \midrule
SPECTRA &  2 & 50 & 45 &  10 & 60\\
HardKuma & 1  & 25 & 15 & 6 & 30 \\
SFE & 1 & 25 & 15 &  6 & 30\\
Sparsemax & 1 & 20 & 15 & 6  & 30 \\ \bottomrule
\end{tabular}
\caption{\label{tab:comp_times_highlights}
Average training and validation time per epoch respective to some strategies used in the paper for each dataset used for highlights extraction.
}
\end{table}

\renewcommand{\arraystretch}{1}
\begin{table}[h!]
\footnotesize
\centering
\begin{tabular}{
>{\arraybackslash}m{3.2cm}
>{\raggedleft\arraybackslash}m{1.6cm}
>{\raggedleft\arraybackslash}m{1.6cm}}
\toprule
Strategy & SNLI & MNLI \\ \midrule
\multicolumn{3}{l}{\textbf{\textit{Training Time}} (in minutes)} \\ \midrule
ESIM & 26  & 23  \\
M:XorAtMostOne & 53   & 60   \\
 M:AtMostOne2 & 53  & 60  \\
 M:Budget & 56  & 65  \\
Gumbel & 28  &  23 \\ \midrule \midrule
\multicolumn{3}{l}{\textbf{\textit{Validation Time}} (in seconds)} \\ \midrule
ESIM & 10 & 10\\
M:XorAtMostOne &  65 &  60\\
M:AtMostOne2 &  65 & 70  \\
M:Budget & 110 & 115\\
Gumbel & 10 & 10\\
\bottomrule

\end{tabular}
\caption{\label{tab:comp_times_matchings}
Average training and validation time per epoch for each dataset and each strategy used for matchings extraction.
}
\end{table}

\section{SPECTRA performance for varying rationale length/alignment budget}
To analyse how SPECTRA fares for different sparsity levels, we ran experiments for highlights extraction and matchings extraction (see Table \ref{tab:budget_varying_highlights} and Table \ref{tab:budget_varying_matchings}) for different budget values.

\renewcommand{\arraystretch}{1}
\begin{table}[h!]
\footnotesize
\centering
\begin{tabular}{
>{\arraybackslash}m{0.2cm}
>{\raggedleft\arraybackslash}m{0.78cm}
>{\raggedleft\arraybackslash}m{1.35cm} 
>{\raggedleft\arraybackslash}m{1.1cm}
>{\raggedleft\arraybackslash}m{0.85cm}
>{\raggedleft\arraybackslash}m{1.1cm}}
\toprule
$B$ & SST $\uparrow$ & AgNews $\uparrow$ & IMDB $\uparrow$ & Beer $\downarrow$ & Hotels $\uparrow$\\ \midrule
$2$ &  \multicolumn{1}{r}{0.578} & \multicolumn{1}{r}{0.870} & \multicolumn{1}{r}{0.891} & \multicolumn{1}{r}{0.020} & 0.875 \\
$5$ & \multicolumn{1}{r}{0.726} & \multicolumn{1}{r}{0.903}& \multicolumn{1}{r}{0.900} & \multicolumn{1}{r}{0.019} & 0.867 \\
$10$ & \multicolumn{1}{r}{0.799} & \multicolumn{1}{r}{0.915} & \multicolumn{1}{r}{0.891} & \multicolumn{1}{r}{0.016} & 0.890 \\
$15$ & \multicolumn{1}{r}{0.798} & \multicolumn{1}{r}{0.916} & \multicolumn{1}{r}{0.893} & \multicolumn{1}{r}{0.016} & 0.906 \\
$20$ & \multicolumn{1}{r}{0.803} & \multicolumn{1}{r}{0.918} & \multicolumn{1}{r}{0.891} & \multicolumn{1}{r}{0.017} & 0.895  \\
\bottomrule

\end{tabular}
\caption{\label{tab:budget_varying_highlights}
Model predictive performances across datasets and different budget values $B$ for the SPECTRA method for highlights extraction. We report $F_1$ scores on test sets for all datasets but Beer where we report MSE. These results are respective to one random seed.}
\end{table}

\renewcommand{\arraystretch}{1}
\begin{table}[h!]
\footnotesize
\centering
\begin{tabular}{
>{\arraybackslash}m{0.2\columnwidth}
>{\raggedleft\arraybackslash}m{0.15\columnwidth}
>{\raggedleft\arraybackslash}m{0.15\columnwidth}}
\toprule
Budget $B$ & SNLI $\uparrow$ & MNLI $\uparrow$ \\ \midrule
$4$ & \multicolumn{1}{r}{0.857} & \multicolumn{1}{r}{0.731}\\
$5$ & \multicolumn{1}{r}{0.851} & \multicolumn{1}{r}{0.731} \\
$6$ & \multicolumn{1}{r}{0.852} & \multicolumn{1}{r}{0.755} \\
\bottomrule
\end{tabular}
\caption{\label{tab:budget_varying_matchings}
Model predictive performances across datasets and different budget values for the SPECTRA method for matchings extraction. We report $F_1$ scores on test sets for all datasets. These results are respective to one random seed.}
\end{table}

\section{Highlights Extracted with SPECTRA}
\label{app:example_highlights_rationales}

Figure \ref{fig:example_rationales} shows examples
of highlights extracted by SPECTRA model on the AgNews and Beer dataset. Interestingly, when compared to human annotations on the Beer dataset, we notice that SPECTRA usually disregards highlighting stopwords. While these explanations do not lose relevant meaning when compared to the human explanations, this ultimately slightly hinders the performance on the matching with human annotations.

\begin{figure}[h!]
\centering
  \includegraphics[width=\columnwidth]{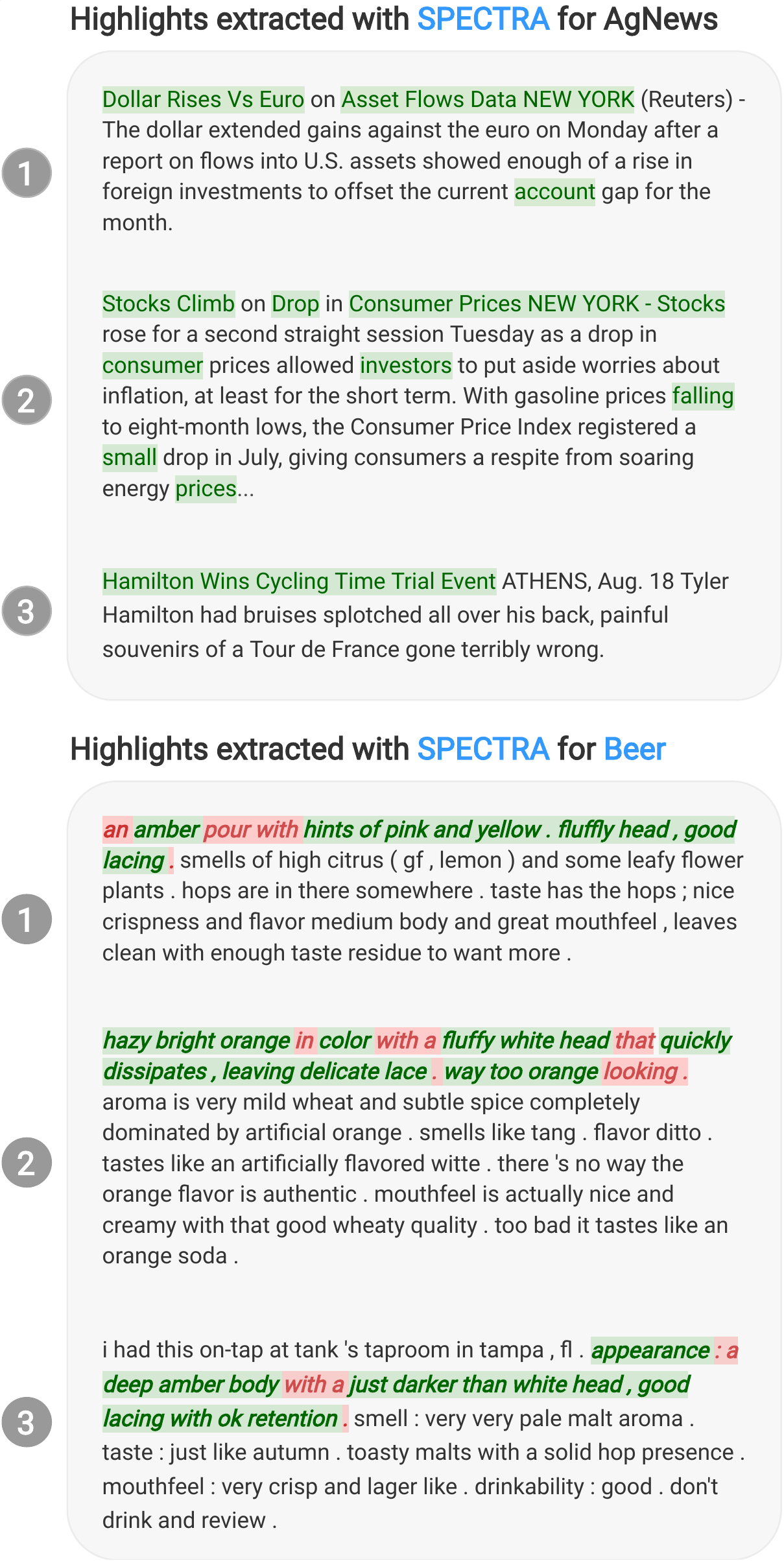}
  \caption{Examples of extracted highlights (green shaded input tokens) with SPECTRA for AgNews and Beer documents. For the rationales with Beer, we show the human annotations in bold and italic (we shade in red the mismatches with the human annotations).}
  \label{fig:example_rationales}
\end{figure}

\section{Matchings Extracted with SPECTRA}
\label{app:example_matchings_rationales}

\paragraph{Synthetic Matchings.} In Figure \ref{fig:synthetic_matchings} we show the extracted matchings with the three different SPECTRA factors that we used in the paper for a synthetic score matrix. The \textsf{M:XOR-AtMostOne} factor constraints the alignment matrix $\bm{Z} \in \mathbb{R}^{L_P \times L_H}$ to be such that for each line $i$ in $\bm{Z}$, we have $\sum_{n=1}^{L_H} z_{in} = 1$. For \textsf{M:AtMostOne2} we have that for each line $i$ in $\bm{Z}$, $\sum_{n=1}^{L_H} z_{in} \leq 1$. And, finally, the more constrained version of \textsf{M:Budget} is such that for each line $i$ in $\bm{Z}$, we have $\sum_{n=1}^{L_H} z_{in} \leq B$, in which $B$ is the Budget value.

\begin{figure}[h!]
\centering
  \includegraphics[width=\columnwidth]{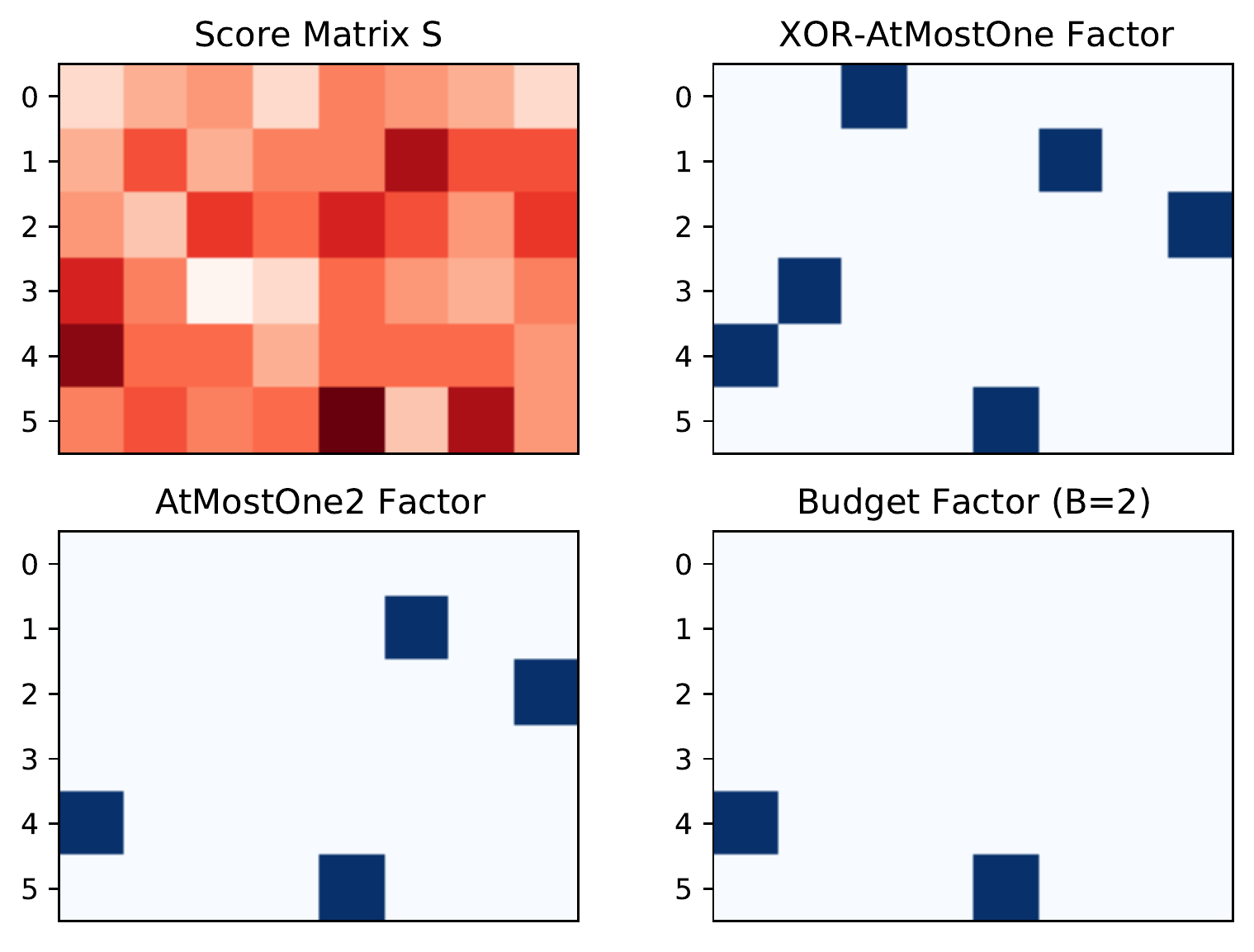}
  \caption{Extracted matchings with the SPECTRA strategies for a synthetic $6 \times 8$ matrix.}
  \label{fig:synthetic_matchings}
\end{figure}

\paragraph{Examples extracted from SNLI.} We show in Figures \ref{fig:example_matchings_xoratmostone}, \ref{fig:example_matchings_atmostone2} and \ref{fig:example_matchings_budget} some examples of matchings extracted with the three different SPECTRA strategies on the SNLI dataset. For these examples all non-null entries in $\bm{Z}$ have value $1$.

\paragraph{Examples extracted from HANS.} We show in Figure \ref{fig:example_matchings_hans} examples of matchings extracted with SPECTRA for the model trained on MNLI augmented with HANS-like examples (Augmented). For all these examples, the original MNLI model without augmentation (Vanilla) classified the examples as entailment, whereas the Augmented model correctly classified them as non-entailment. Interestingly, the obtained matchings highlight the use of the heuristics that HANS aims to target. However, the Augmented model is able to process the information from the matchings in such a way that it correctly classifies most non-entailment examples (see Table \ref{tab:results_HANS}).

\begin{figure}[ht]
\centering
  \includegraphics[width=\columnwidth]{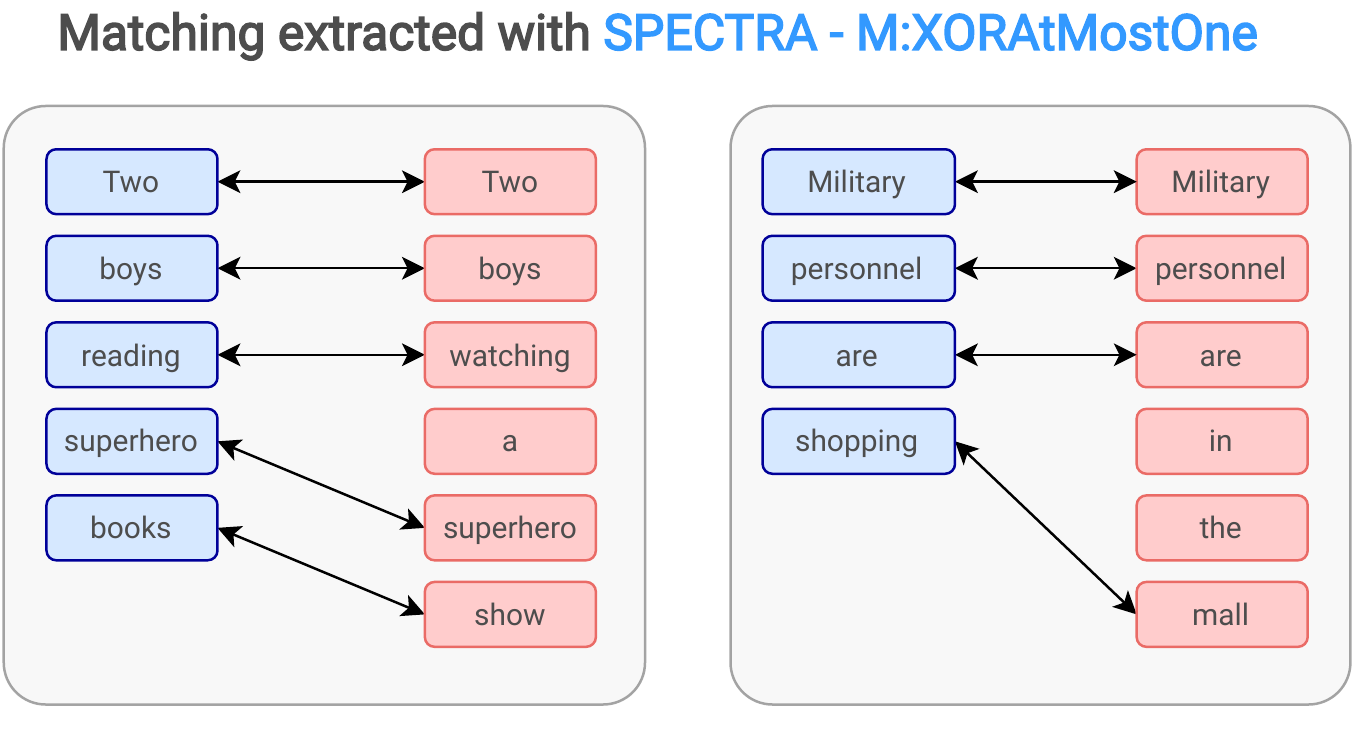}
  \caption{Examples of extracted matchings with \textsf{M:XORAtMostOne}. The premise is shown on the left and the hypothesis is shown on the right.}
  \label{fig:example_matchings_xoratmostone}
\end{figure}
\vfill

\begin{figure}[h!]
\centering
  \includegraphics[width=\columnwidth]{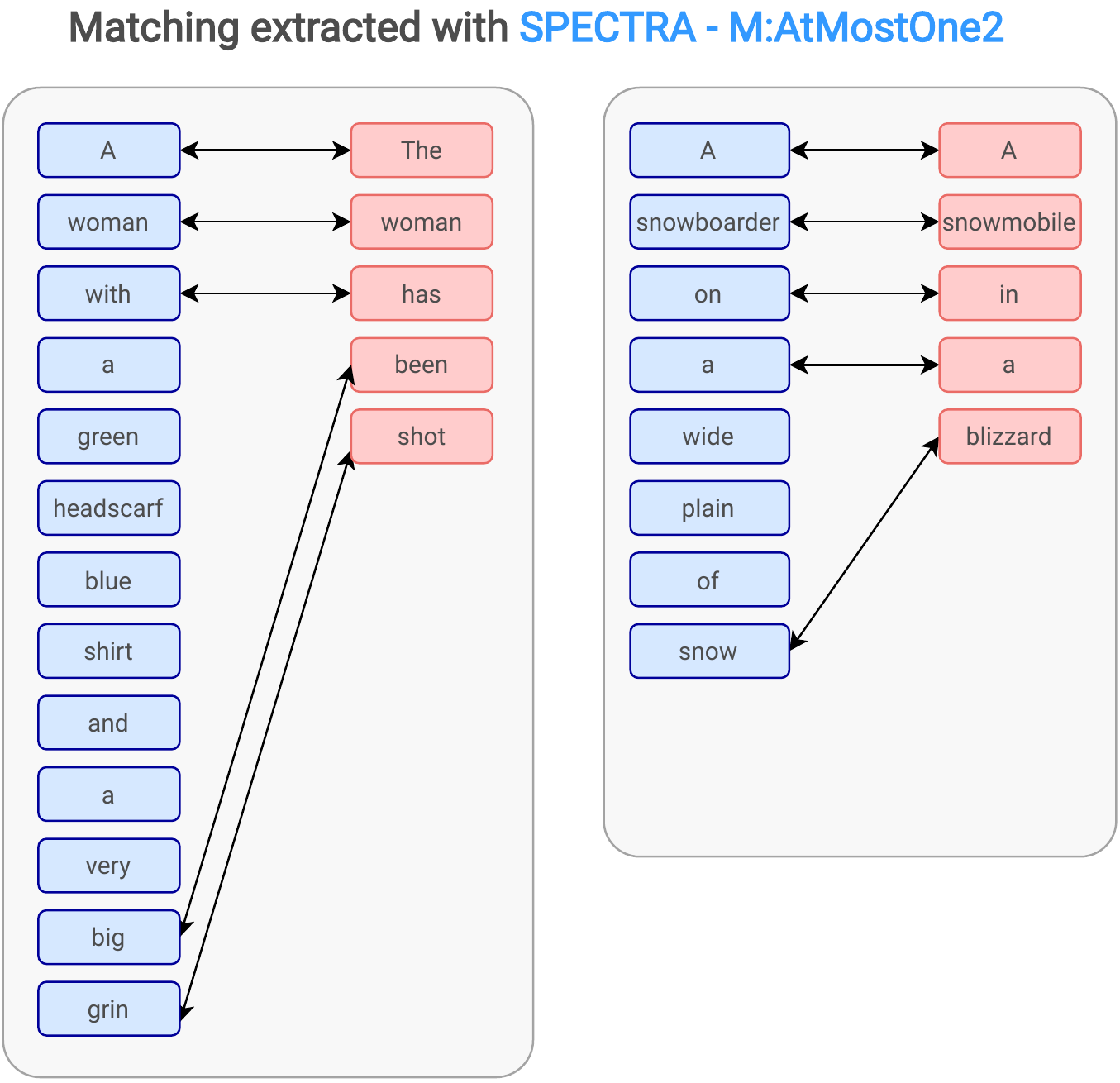}
  \caption{Examples of extracted matchings with \textsf{M:AtMostOne2}. The premise is shown on the left and the hypothesis is shown on the right.}
  \label{fig:example_matchings_atmostone2}
\end{figure}

\vfill
\newpage

\begin{figure}[h]
\centering
  \includegraphics[width=\columnwidth]{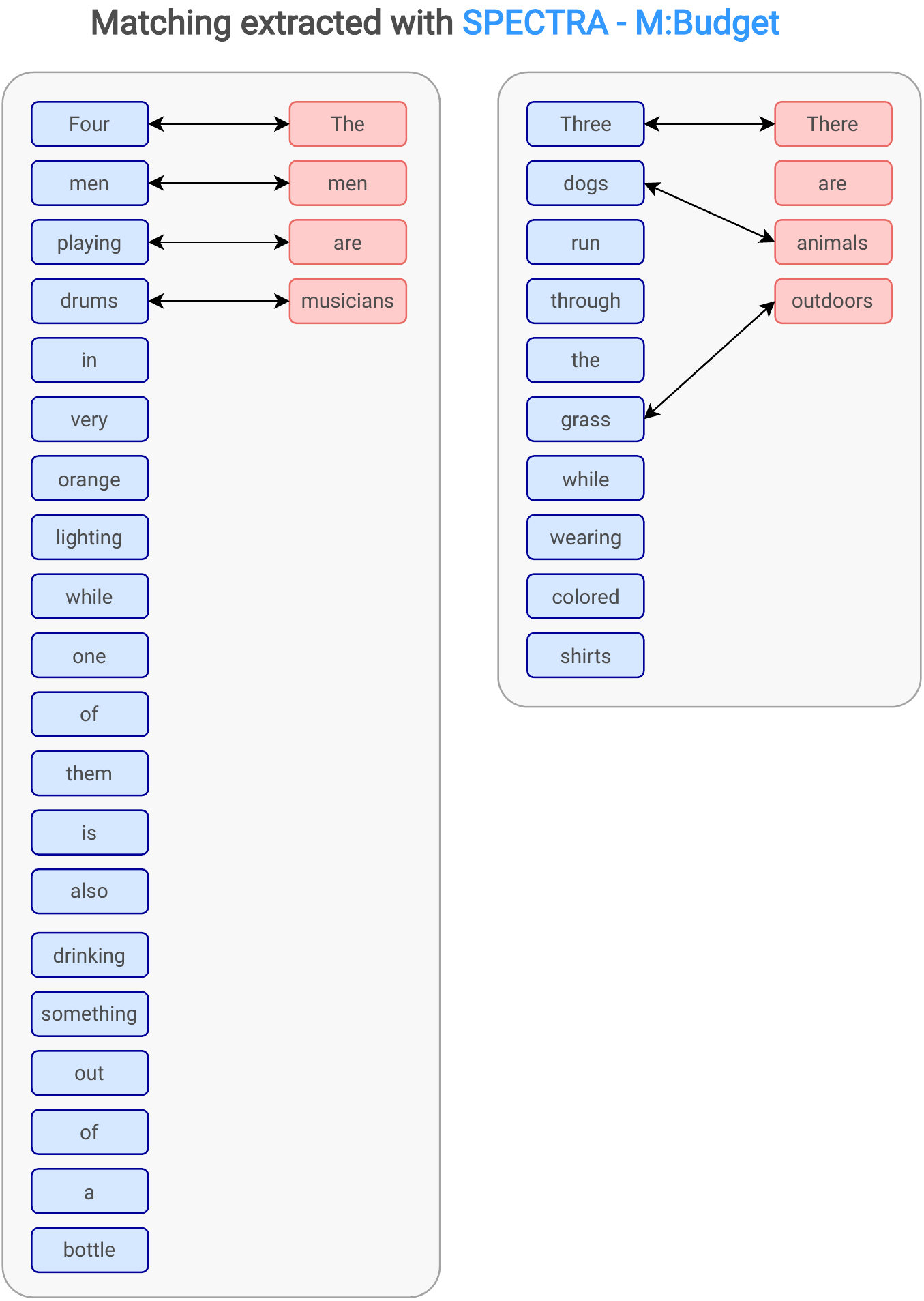}
  \caption{Examples of extracted matchings with \textsf{M:Budget} -- the Budget value is set to 4. The premise is shown on the left and the hypothesis is shown on the right.}
  \label{fig:example_matchings_budget}
\end{figure}

\begin{figure}[t]
\centering
  \includegraphics[width=.8\columnwidth]{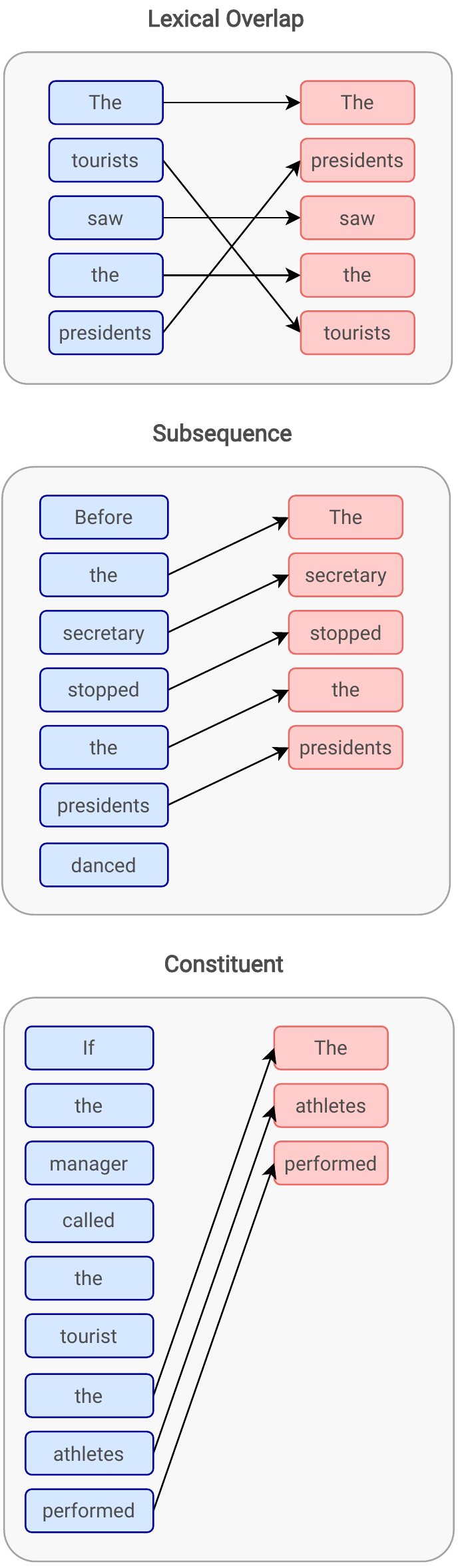}
  \caption{Examples of extracted matchings with SPECTRA (Augmented) that highlight the three linguistic heuristics of HANS: lexical overlap, constituent and subsequence heuristics. The premise is shown on the left and the hypothesis is shown on the right.}
  \label{fig:example_matchings_hans}
\end{figure}

\vfill

\end{document}